\documentclass{article}
\usepackage{ijcai24}

\usepackage{times}
\usepackage{soul}
\usepackage{url}
\usepackage[hidelinks]{hyperref}
\usepackage[utf8]{inputenc}
\usepackage[small]{caption}
\usepackage{graphicx}
\usepackage{amsmath}
\usepackage{amsthm}
\usepackage{booktabs}
\usepackage{algorithm}
\usepackage{algorithmic}
\usepackage[switch]{lineno}

\newtheorem{example}{Example}
\newtheorem{theorem}{Theorem}

\title{Learning Big Logical Rules by Joining Small Rules}
\author{
  C\'{e}line Hocquette\textsuperscript{\rm 1}, Andreas Niskanen\textsuperscript{\rm 2},
  Rolf Morel\textsuperscript{\rm 1}, 
    Matti J\"{a}rvisalo\textsuperscript{\rm 2},
    Andrew Cropper\textsuperscript{\rm 1}\\
    \textsuperscript{\rm 1}University of Oxford\\
    \textsuperscript{\rm 2}University of Helsinki\\
    celine.hocquette@cs.ox.ac.uk 
}

\usepackage[utf8]{inputenc}

\usepackage{xcolor}
\usepackage{booktabs}
\usepackage{listings}
\usepackage{inconsolata}
\usepackage{tikz}
\usepackage{pgfplots}
\usepackage{amsmath}
\usepackage{microtype}
\usepackage{algorithm}
\usepackage{amssymb}
\usepackage[titletoc]{appendix}
 \usepackage{cancel}
 \usepackage{multirow}
\usepackage{subfig}
\newcommand{\name}{\textsc{Joiner}}

\newcommand{\combo}{\textsc{Combo}}

\newcommand{\ilasp}{\textsc{ILASP}}
\newcommand{\ale}{\textsc{Aleph}}

\newcommand{\aspal}{\textsc{Aspal}}

\theoremstyle{definition}
\newtheorem{definition}{Definition}
\newtheorem{myexample}{Example}
\newtheorem{lemma}{Lemma}
\newtheorem{assumption}{Assumption}
\newtheorem{corollary}{Corollary}
\newtheorem{proposition}{Proposition}

\lstnewenvironment{myalgorithm}[1][] %defines the algorithm listing environment
{
    \lstset{ %this is the stype
        basicstyle=\normalfont\ttfamily,
        showspaces=false,               % show spaces adding particular underscores
        showstringspaces=false,         % underline spaces within strings
        mathescape=true,
        numbers=left,
        escapeinside={*}{*},
        columns=flexible,
        numbersep=2pt,        % default 10pt
        keywordstyle=\bfseries,
        keywords={,and, return, not, def, in, if, elif, else, for, foreach, while, continue,break,}
        numbers=left,
        xleftmargin=6pt
        #1 % this is to add specific settings to an usage of this environment (for instnce, the caption and referable label)
    }
}
{}

\usepackage{pgfkeys}
    \newenvironment{customlegend}[1][]{%
        \begingroup
        \csname pgfplots@init@cleared@structures\endcsname
        \pgfplotsset{#1}%
    }{%
        \csname pgfplots@createlegend\endcsname
        \endgroup
    }%
    \def\addlegendimage{\scriptsize\csname pgfplots@addlegendimage\endcsname}
\usepackage{comment}

\usepackage{pgfplots}
\usepgfplotslibrary{colorbrewer}
\pgfplotsset{compat = 1.15, cycle list/Set1-8} 
\usetikzlibrary{pgfplots.statistics, pgfplots.colorbrewer} 
\usepackage{pgfplotstable}
\usepackage{filecontents}

\begin{document}

\maketitle

\begin{abstract}
A major challenge in inductive logic programming is learning big rules.
To address this challenge, we introduce an approach where we join small rules to learn big rules. 
We implement our approach in a constraint-driven system and use constraint solvers to efficiently join rules.
Our experiments on many domains, including game playing and drug design, show that our approach can (i) learn rules with more than 100 literals, and (ii) drastically outperform existing approaches in terms of predictive accuracies.
\end{abstract}

\section{Introduction}
\label{introduction}
\begin{figure}[!ht]
\footnotesize
\centering
\fbox{
\includegraphics[height=1.2cm]{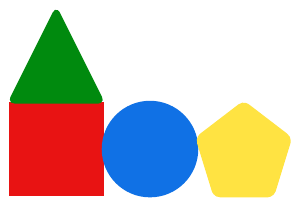}
}
\fbox{
\includegraphics[height=1.2cm]{./figures/intro/zendo5.pdf}
}
\caption{
Two positive Zendo examples.
}
\label{fig:posexamples}
\end{figure}

Zendo is an inductive reasoning game. 
One player, the \emph{teacher}, creates a secret rule that describes structures.
The other players, the \emph{students}, try to discover the secret rule by building structures.
The teacher marks whether structures follow or break the rule. 
The first student to correctly guess the rule wins.
For instance, for the positive examples shown in Figure~\ref{fig:posexamples}, a possible rule is \emph{``there is a blue piece''}.

\noindent
To use machine learning to play Zendo, we need to learn explainable rules from a small number of examples.
Although crucial for many real-world problems, many machine-learning approaches struggle with this type of learning \cite{ilp30}.
Inductive logic programming (ILP) \cite{mugg:ilp} is a form of machine learning that can learn explainable rules from a small number of examples. 
For instance, for the examples in Figure \ref{fig:posexamples}, an ILP system could learn the rule:
\[
    h_1 = 
    \begin{array}{l}
    \left\{
    \begin{array}{l}
\emph{zendo(S) $\leftarrow$ piece(S,B), blue(B)}\\
    \end{array}
    \right\}
    \end{array}
\]

\noindent
This rule says that the relation \emph{zendo} holds for the structure $S$ when there is a blue piece $B$.

\begin{figure}[ht]
\footnotesize
\centering
\fbox{
\includegraphics[height=1.2cm]{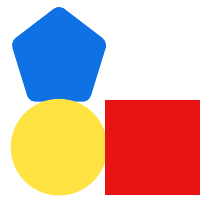}
}
\fbox{
\includegraphics[height=1.2cm]{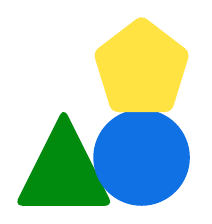}
}
\fbox{
\includegraphics[height=1.2cm]{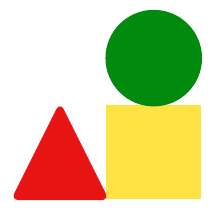}
}
\caption{
Three negative Zendo examples.
}
\label{fig:negexamples}
\end{figure}

% \noindent
Suppose we also have the three negative examples shown in Figure \ref{fig:negexamples}.
Our previous rule incorrectly entails the first and second negative examples, as they have a blue piece.
To entail all the positive and no negative examples, we need a bigger rule, such as:
\[
    h_2 = 
    \begin{array}{l}
    \left\{
    \begin{array}{l}
\emph{zendo(S) $\leftarrow$ piece(S,B), blue(B),}\\
\hspace{50pt}\emph{piece(S,R), red(R),}\\
\hspace{50pt}\emph{piece(S,G), green(G)}\\
    \end{array}
    \right\}
    \end{array}
\]
\noindent
% This rule says that the relation \emph{zendo} holds for a structure when there is a blue, a red, a green, and a yellow piece.
This rule says that the relation \emph{zendo} holds for a structure when there is a blue piece, a red piece, and a green piece.

% \noindent
Most ILP approaches can learn small rules, such as $h_1$.
However, many struggle to learn bigger rules, such as $h_2$\footnote{A rule of size 7 is not especially big but for readability we do not use a bigger rule in this example.
In our experiments, we show we can learn similar rules with over 100 literals.
}.
% \emph{rule selection} 
This limitation noticeably holds for approaches which precompute all possible rules and search for a subset of them \cite{aspal,ilasp,hexmil,difflog,prosynth,apperception,aspsynth}.
As they precompute all rules, these approaches struggle to learn rules with more than a few literals. 

To address this limitation, we introduce an approach that learns big rules by joining small rules.
The idea is to first find small rules where each rule entails some positive and some negative examples.
We then search for conjunctions of these small rules such that each conjunction entails at least one positive example but no negative examples.

To illustrate our idea, consider our Zendo scenario.
We first search for rules that entail at least one positive example, such as:
% Rather than directly trying to learn $h_2$, we instead first find small rules, 
% such as:
\[
    \begin{array}{l}
    r_1 = \left\{
    \begin{array}{l}
\emph{zendo$_1$(S) $\leftarrow$ piece(S,B), blue(B)}
    \end{array}
    \right\}\\
    r_2 = \left\{
    \begin{array}{l}
\emph{zendo$_2$(S) $\leftarrow$ piece(S,R), red(R)}
    \end{array}
    \right\}\\
    r_3 = \left\{
    \begin{array}{l}
\emph{zendo$_3$(S) $\leftarrow$ piece(S,G), green(G)}
    \end{array}
    \right\}\\
    r_4 = \left\{
    \begin{array}{l}
\emph{zendo$_4$(S) $\leftarrow$ piece(S,Y), yellow(Y)}
    \end{array}
    \right\}
    \end{array}
\]

\noindent
% Each rule entails at least one positive example but also at least one negative example.
Each of these rules entails at least one negative example.
Therefore, we search for a subset of these rules where the intersection of the logical consequences of the subset entails at least one positive example and no negative examples.
% conjunction 
The set of rules $\{r_1, r_2, r_3\}$ satisfies these criteria.
We therefore form a hypothesis from the conjunction of these rules:
\[
    \begin{array}{l}
    h_3 = \left\{
    \begin{array}{l}
\emph{zendo(S) $\leftarrow$ zendo$_1$(S), zendo$_2$(S), zendo$_3$(S)}
    \end{array}
    \right\}
    \end{array}
\]
The hypothesis $h_3$ entails all the positive but none of the negative examples and has the same logical consequences (restricted to \emph{zendo/1} atoms) as $h_2$.
% of joining small rules to learn large rules 

% \noindent
The main benefit of our approach is that we can learn rules with over 100 literals, which existing approaches cannot.
% , sometimes learning rules with over 100 literals.
Our approach works because we decompose a learning task into smaller tasks that can be solved separately. 
For instance, instead of directly searching for a rule of size 7 to learn $h_2$, we search for rules of size 3 ($r_1$ to $r_3$) and try to join them to learn $h_3$.
As the search complexity of ILP is usually exponential in the size of the program to be learned, this reduction can substantially improve learning performance.
Moreover, because we can join small rules to learn big rules of a certain syntactic form, we can eliminate \emph{splittable} rules from the search space.
We formally define a splittable rule in Section \ref{join}, but informally the body of a splittable rule can be decomposed into independent subsets, such as the body of $h_2$ in our Zendo example.

% many rules from the search space.
% Specifically, we

% For instance, in our \emph{zendo} domain from our experiments, there are 176,453 possible rules, but only 52,390 are non-splittable.
% \ac{I still need to improve this paragraph}

To explore our idea, we build on \emph{learning from failures} (LFF) \cite{popper}, a constraint-driven ILP approach. 
We extend the LFF system \combo{} \cite{combo} with a \emph{join stage} to learn programs with big rules. We develop a Boolean satisfiability (SAT)~\cite{sat-handbook} approach to find conjunctions in the join stage.
% and use SAT solvers \cite{PySAT} to efficiently find conjunctions.
% with subset-maximal coverage.
We call our implementation \name{}.

\paragraph{Novelty and Contributions.}
Our main contribution is the idea of joining small rules to learn big rules which, as our experiments on many diverse domains show, can improve predictive accuracies.
Overall, our main contributions are:
\begin{enumerate}
\item We introduce an approach which joins small rules to learn big rules.
\item We implement our approach in \name, which learns optimal (textually minimal) and recursive programs. 
We prove the correctness of \name{} (Theorem \ref{prop:correct}).
\item We experimentally show on several domains, including game playing, drug design, and string classification, that our approach can (i) learn rules with more than 100 literals, and (ii) drastically outperform existing approaches in terms of predictive accuracy.
\end{enumerate}

\section{Related Work}
\label{sec:related}

\textbf{Program synthesis.}
Several approaches build a program one token at a time using an LSTM \cite{devlin2017robustfill,bunel2018leveraging}. \textsc{CrossBeam} \cite{crossbeam} uses a neural model to generate programs as the compositions of seen subprograms.
\textsc{CrossBeam} is not guaranteed to learn a solution if one exists and can only use unary and binary relations. 
By contrast, \name{} can use relations of any arity.

\textbf{Rule mining.} 
\textsc{AMIE+} \cite{amie+} is a prominent rule mining approach.
In contrast to \name{}, \textsc{AMIE+} can only use unary and binary relations and struggles to learn rules with more than 4 literals.
% By contrast, \name{} can learn rules with over 100 literals.

\textbf{ILP.}
Top-down ILP systems \cite{foil,tilde} specialise rules with refinement operators \cite{shapiro:thesis}.
% These approaches need a user-provided parameter to restrict the maximum size of a rule.
% By contrast, we do not restrict the size of rules in our join stage.
% In addition,
Because they learn a single rule at a time and add a single literal at each step, these systems struggle to learn recursive and optimal programs.
% and perform predicate invention.
Recent approaches overcome these issues by formulating the search as a rule selection problem \cite{aspal,dilp,hexmil,difflog,prosynth,apperception,aspsynth}. 
These approaches precompute all possible rules (up to a maximum rule size) and thus struggle to learn rules with more than a few literals.
By contrast, we avoid enumeration and use constraints to soundly prune rules.
Moreover, our join stage allows us to learn rules with more than 100 literals.

\textbf{Many rules.} 
\combo{} \cite{combo} searches for a disjunction of small programs that entails all the positive examples. 
\combo{} learns optimal and recursive programs and large programs with many rules.
However, it struggles to learn rules with more than 6 literals.
Our approach builds on \combo{} and can learn rules with over 100 literals.

\textbf{Big rules.}
Inverse entailment approaches \cite{progol,aleph} can learn big rules by returning bottom clauses. 
However, these approaches struggle to learn optimal and recursive programs and tend to overfit.
Ferilli \shortcite{pi_spec} specialises rules in a theory revision task. 
This approach can learn negated conjunctions but only Datalog preconditions and not recursive programs. 
\textsc{Brute} \cite{brute} can learn recursive programs with hundreds of literals.
However, in contrast to our approach, \textsc{Brute} needs a user-provided domain-specific loss function, does not learn optimal programs, and can only use binary relations.

\textbf{Splitting rules.}
% \cite{costa2000note}
Costa et al. \shortcite{DBLP:journals/jmlr/CostaSCBDJSVL03} split rules into conjunctions of independent goals that can be executed separately to avoid unnecessary backtracking and thus to improve execution times. 
By contrast, we split rules to reduce search complexity.
% Costa et al. allow variable sharing between joined rules, whereas we prevent joined rules from sharing body-only variables.
Costa et al. allow joined rules to share variables, whereas we prevent joined rules from sharing body-only variables.

\section{Problem Setting}
We now describe our problem setting. We assume familiarity with logic programming \cite{lloyd:book} but have stated relevant notation and definitions in the appendix. % A constraint is a Horn clause without a positive literal.

% \subsection{Learning From Failures}
We use the learning from failures (LFF) \cite{popper} setting. We restate some key definitions \cite{combo}. 
A \emph{hypothesis} is a set of definite clauses with the least Herbrand model semantics. 
We use the term \emph{program} interchangeably with the term hypothesis. 
A \emph{hypothesis space} $\cal{H}$ is a set of hypotheses. 
LFF assumes a language $\cal{L}$ that defines hypotheses. 
% For instance, consider a meta-language with two literals \emph{h\_lit/3} and \emph{b\_lit/3} representing \emph{head} and \emph{body} literals respectively.
% In this meta-language, the hypothesis \emph{zendo(A) $\leftarrow$ piece(A,B), blue(B)} is represented as the set of literals \emph{\{h\_lit(0,zendo,(0)), b\_lit(0,piece,(0,1)), b\_lit(0,blue,(1))\}}.
% The first argument of each of these literals is the rule index, the second is the predicate symbol, and the third is the literal variables (\emph{0} represents the variable \emph{A}, \emph{1} represents \emph{B}, ...).
A LFF learner uses hypothesis constraints to restrict the hypothesis space. 
A \emph{hypothesis constraint} is a constraint (a headless rule) expressed in $\mathcal{L}$.
% Given a set of hypotheses constraints $C$, 
A hypothesis $h$ is consistent with a set of constraints $C$ if, when written in ${\cal{L}}$, $h$ does not violate any constraint in $C$. We call ${\cal{H}}_{C}$ the subset of $\cal{H}$ consistent with $C$. 
We define a LFF input and a solution:
\begin{definition}[\textbf{LFF input}]
A \emph{LFF input} is a tuple $(E, B, \mathcal{H}, C)$ where $E=(E^{+}, E^{-})$ is a pair of sets of ground atoms denoting positive ($E^+$) and negative ($E^-$) examples, $B$ is a definite program denoting background knowledge,
$\mathcal{H}$ is a hypothesis space, and $C$ is a set of hypothesis constraints.
\end{definition}
\noindent
%We define a LFF solution:
\begin{definition}[\textbf{Solution}]
\noindent
For a LFF input $(E, B, \mathcal{H}, C)$, where $E=(E^{+}, E^{-})$, a hypothesis $h \in \mathcal{H}_{C}$ is a \emph{solution} when $h$ entails every example in $E^+$ ($\forall e \in E^+, \; B \cup h \models e$) and no example in $E^-$ ($\forall e \in E^-, \; B \cup h \not\models e$).
\label{def:solution}
\end{definition}

\noindent
Let $cost : \mathcal{H} \mapsto \mathbb{N}$ be an arbitrary cost function that measures the cost of a hypothesis.
We define an \emph{optimal} solution:

\begin{definition}[\textbf{Optimal solution}]
\label{def:opthyp}
For a LFF input $(E, B, \mathcal{H}, C)$, a hypothesis $h \in \mathcal{H}_{C}$ is \emph{optimal} when (i) $h$ is a solution, and (ii) $\forall h' \in \mathcal{H}_{C}$, where $h'$ is a solution, $cost(h) \leq cost(h')$.
\end{definition}
\noindent
%In this paper,
Our cost function is the number of literals in a hypothesis.

A hypothesis which is not a solution is a \emph{failure}. 
% A LFF learner identifies constraints from failures to restrict the hypothesis space. For instance, if a hypothesis does not entail any positive example, a \emph{specialisation} constraint prunes its specialisations, as they will also not entail any positive examples.
% \noindent
For a hypothesis $h$, the number of true positives ($tp$) and false negatives ($fn$) is the number of positive examples entailed and not entailed by $h$ respectively.
% We define the number of true negatives ($tn$) and false positives ($fp$) as the number of negative examples respectively uncovered and covered by $h$.
The number of false positives ($fp$) is the number of negative examples entailed by $h$.

\section{Algorithm}
\label{join}

To describe \name{}, we first describe \combo{} \cite{combo}, which we build on.

\paragraph{\combo{}.} 
\combo{} takes as input background knowledge, positive and negative training examples, and a maximum hypothesis size.
\combo{} builds a constraint satisfaction problem (CSP) program $\mathcal{C}$, where each model of $\mathcal{C}$ corresponds to a hypothesis (a definite program).
In the \emph{generate stage}, \combo{} searches for a model of $\mathcal{C}$ for increasing hypothesis sizes.
If no model is found, \combo{} increments the hypothesis size and resumes the search.
If there is a model, \combo{} converts it to a hypothesis $h$.
In the \emph{test stage}, \combo{} uses Prolog to test $h$ on the training examples.
If $h$ is a solution, \combo{} returns it.
Otherwise, if $h$ entails at least one positive example and no negative examples, \combo{} saves $h$ as a \emph{promising program}.
In the \emph{combine stage}, \combo{} searches for a combination (a union) of promising programs that entails all the positive examples and is minimal in size.
If there is a combination, \combo{} saves it as the best combination so far and updates the maximum hypothesis size.
In the \emph{constrain stage}, \combo{} uses $h$ to build constraints and adds them to $\mathcal{C}$ to prune models and thus prune the hypothesis space.
For instance, if $h$ does not entail any positive example, \combo{} adds a constraint to eliminate its specialisations as they are guaranteed to not entail any positive example.
\combo{} repeats this loop until it finds an optimal (textually minimal) solution or exhausts the models of $\mathcal{C}$.

\subsection{Joiner}
Algorithm \ref{alg:metaspec} shows our \name{} algorithm.
 % which builds on \combo{}.
% It is similar to \combo{} but has three key differences.
\name{} builds on \combo{} and uses a generate, test, join, combine, and constrain loop to find an optimal solution (Definition \ref{def:opthyp}).
\name{} differs from \combo{} by (i) eliminating splittable programs in the generate stage (line 5), (ii) using a join stage to build big rules from small rules and saving them as promising programs (line 16), and (iii) using different constraints (line 20).
We describe these differences in turn.
\begin{algorithm}[t]
\footnotesize
{
\begin{myalgorithm}[]
def $\name$(bk, $E^+$, $E^-$, maxsize):
  cons, to_join, to_combine = {}, {}, {}
  bestsol, k = None, 1
  while k $\leq$ maxsize:
    h = generate_non_splittable(cons, k)
    if h == UNSAT:
      k += 1
      continue
    tp, fn, fp = test($E^+$, $E^-$, bk, h)
    if fn == 0 and fp == 0:
      return h
    elif tp > 0 and fp == 0:
      to_combine += {h}
    elif tp > 0 and fp > 0:
      to_join += {h}
    to_combine += join(to_join, bestsol,$E^+$, $E^-$, k)
    combi = combine(to_combine, maxsize, bk, $E^-$)
    if combi != None:
      bestsol, maxsize = combi, size(bestsol)-1
    cons += constrain(h, tp, fp)
  return bestsol
\end{myalgorithm}
\caption{
\name{}
}
\label{alg:metaspec}
}
\end{algorithm}

\subsection{Generate}
In our generate stage, we eliminate \emph{splittable} programs because we can build them in the join stage.
We define a splittable rule:
\begin{definition}[\textbf{Splittable rule}]
A rule is splittable if its body literals can be partitioned into two non-empty sets such that the body-only variables (a variable in the body of a rule but not the head) in the literals of one set are disjoint from the body-only variables in the literals of the other set.
A rule is non-splittable if it is not splittable.
\label{def:splittablerule}
\end{definition}
\begin{myexample}\textbf{(Splittable rule)}
Consider the rule:
\[
\begin{array}{l}
% r_1 = 
\left\{
\begin{array}{l}
\emph{zendo(S) $\leftarrow$ piece(S,R), red(R), piece(S,B), blue(B)}\\
\end{array}
\right\}
\end{array}
\]
This rule is splittable because its body literals can be partitioned into two sets \{\emph{piece(S,R), red(R)}\} and \{\emph{piece(S,B), blue(B)}\}, with body-only variables \emph{\{R\}} and \emph{\{B\}} respectively.
% Consider the rule:
% % \ac{make into a non-splittable example}
% \[
% \begin{array}{l}
% r_2 = \left\{
% \begin{array}{l}
% \emph{zendo(S) $\leftarrow$ closed(S), piece(S,B), blue(B)}\\
% \end{array}
% \right\}
% \end{array}
% \]
% This rule is splittable because its body literals can be partitioned into two sets \{\emph{closed(S)}\} and \{\emph{piece(S,B), blue(B)}\}, where the body-only variables of each set are \{\} and \{B\} respectively.
\end{myexample}

\begin{myexample}\textbf{(Non-splittable rule)}
Consider the rule:
\[
\begin{array}{l}
% r_3 = 
\left\{
\begin{array}{l}
\emph{zendo(S) $\leftarrow$ piece(S,R), red(R), piece(S,B),}\\
\hspace{51pt}\emph{blue(B), contact(R,B)}\\
\end{array}
\right\}
\end{array}
\]
This rule is non-splittable because each body literal contains the body-only variable $R$ or $B$ and one literal contains both.
\end{myexample}

\noindent
We define a splittable program:
\begin{definition}
[\textbf{Splittable program}]
A program is \emph{splittable} if and only if it has exactly one rule and this rule is splittable.
A program is non-splittable when it is not splittable.
\label{def:splittable}
\end{definition}

\noindent
% To eliminate splittable programs in the generate stage, 
We use a constraint to prevent the CSP solver from considering models with splittable programs. 
The appendix includes our encoding of this constraint.
At a high level, we first identify connected body-only variables.
Two body-only variables $A$ and $B$ are connected if they are in the same body literal, or if there exists another body-only variable $C$ such that $A$ and $C$ are connected and $B$ and $C$ are connected. Our constraint prunes programs with a single rule which has (i) two body-only variables that are not connected, or (ii) multiple body literals and at least one body literal without body-only variables.
As our experiments show, eliminating splittable programs can substantially improve learning performance.
% reduce the complexity of the generate stage.

\subsection{Join}
Algorithm \ref{alg:join} shows our join algorithm, which takes as input a set of programs and their coverage ($\mathcal{P}$), where each program entails some positive and some negative examples, the best solution found thus far (\emph{bestsol}), the positive examples ($E^+$), the negative examples ($E^-$), and a maximum conjunction size (\emph{k}).
% explain informally what a conjunction is
It returns subsets of $\mathcal{P}$, where the intersection of the logical consequences of the programs in each subset entails at least one positive example and no negative example.
We call such subsets \emph{conjunctions}. 
% output conjunction entails (i) at least one positive example ($tp > 0$), and (ii) no negative example ($fp = 0$).
% It outputs subsets of these programs where the intersection of the logical consequences of the programs in the subset covers (i) at least one positive example ($tp > 0$), and (ii) no negative example ($fp = 0$).

We define a conjunction:
\begin{definition}[\textbf{Conjunction}]
% Let $C=\{P_1, P_2, ...\}$ be a set of hypotheses with the same head literal $p(a_0, a_1, ...)$.
% Then the conjunction of $S$ is the program $\{P, P'_1, P'_2, ... \}$ where each $P'_i$ is obtained by substituting the head predicate \ac{symbol?} $p$ in $P_i$ with a new predicate symbol $p_i$ and $P = \{p(a_0, a_1, ...) \leftarrow p_0(a_0, a_1, ...), p_1(a_0, a_1, ...), ...\}$. 
A conjunction is a set of programs with the same head literal. 
We call $M(p)$ the least Herbrand model of the logic program $p$. 
The logical consequences of a conjunction $c$ is the intersection of the logical consequences of the programs in it: $M(c) = \cap_{p \in c} M(p)$. 
The cost of a conjunction $c$ is the sum of the cost of the programs in it: $cost(c) = \sum_{p\in c} cost(p)$.

\end{definition}
% \begin{myexample}[\textbf{Conjunction}]
% Let $S$ be the set of programs:
% \[
%     S=\left\{\begin{array}{l}
%     \left\{
%     \begin{array}{l}
% \emph{zendo(S) $\leftarrow$ piece(S,B), blue(B)}
%     \end{array}
%     \right\}\\
%     \left\{
%     \begin{array}{l}
% \emph{zendo(S) $\leftarrow$ piece(S,R), red(R)}
%     \end{array}
%     \right\}
%     \end{array}
%     \right\}
% \]
% The conjunction of $S$ is:
% \[
%     \begin{array}{l}
%     \left\{
%     \begin{array}{l}
% \emph{zendo(S) $\leftarrow$ z1(S), z2(S)}\\
% \emph{z1(S) $\leftarrow$ piece(S,B), blue(B)}\\
% \emph{z2(S) $\leftarrow$ piece(S,R), red(R)}
%     \end{array}
%     \right\}
% \end{array}
% \]

% \begin{figure}
% \begin{footnotesize}
% \begin{tabular}{l}
% \emph{f(S) $\leftarrow$  f$_{1}$(S), f$_{2}$(S)}\\
% \emph{f$_{1}$(S) $\leftarrow$  head(S,H), a(H)}\\
% \emph{f$_{1}$(S) $\leftarrow$  tail(S,T), f$_{1}$(T)}\\
% \emph{f$_{2}$(S) $\leftarrow$  head(S,H), b(H)}\\
% \emph{f$_{2}$(S) $\leftarrow$  tail(S,T), f$_{2}$(T)}\\
% \end{tabular}
% \end{footnotesize}
% % \caption{Example of hypothesis for the \emph{string} domain. This hypothesis states that a string is positive if it contains both the letter \emph{a} followed by the letter \emph{b} and the letter \emph{c} followed by the letter \emph{d}. We vary the number of letters to identify in a string hence the number of body literals.}
% \caption{Example of a recursive conjunction. A string is positive if it contains the letters \emph{a} and \emph{b}.
% }
% \label{hypstring1}
% \end{figure}

\noindent
Our join algorithm has two stages.
We first search for conjunctions that together entail all the positive examples, which allows us to quickly find a solution (Definition \ref{def:solution}).
If we have a solution, we enumerate all remaining conjunctions to guarantee optimality (Definition \ref{def:opthyp}).
In other words, at each call of the join stage (line 16 in Algorithm \ref{alg:metaspec}), we either run the incomplete or the complete join stage (line 3 or 5 in Algorithm \ref{alg:join}).
We describe these two stages.
% Because we assume a solution exists, we are guaranteed to call the complete join stage after some iterations.

\subsubsection{Incomplete Join Stage}
If we do not have a solution, we use a greedy set-covering algorithm to try to cover all the positive examples.
We initially mark each positive example as uncovered (line 8).
We then search for a conjunction that entails the maximum number of uncovered positive examples (line 11).
We repeat this loop until we have covered all the positive examples or there are no more conjunctions. 
This step allows us to first find conjunctions with large coverage and quickly build a solution.
However, this solution may not be optimal.
% , and to quickly find a solution, although not an optimal one. 
% Therefore, in a first step, we enumerate conjunctions by decreasing number of uncovered positive examples covered. 

\subsubsection{Complete Join Stage}
% If we have a solution, the join stage finds all remaining conjunctions. This step ensures we consider all splittable programs and thus ensures optimality.
If we have a solution, we find all remaining conjunctions to ensure we consider all splittable programs and thus ensure optimality.
However, we do not need to find all conjunctions as some cannot be in an optimal solution.
If a conjunction entails a subset of the positive examples entailed by a strictly smaller conjunction then it cannot be in an optimal solution:
% \begin{proposition}[\textbf{Subsumed conjunction}]
\begin{proposition}
Let $c_1$ and $c_2$ be two conjunctions which do not entail any negative examples,
% and entail the positive examples $E^{+}_1$ and $E^{+}_2$ respectively, and
% which entail the positive examples 
$c_1 \models E^{+}_1$,
$c_2 \models E^{+}_2$,
% and $E^{+}_2$ respectively and no negative example, 
% such that
$E^{+}_2 \subseteq E^{+}_1$, 
and $size(c_1) < size(c_2)$. 
Then $c_2$ cannot be in an optimal solution.

\label{prop:subsumed}
\end{proposition}
\noindent
The appendix contains a proof for this result. 
Following this result, our join stage enumerates conjunctions by increasing size. For increasing values of $k$, we search for all \emph{subset-maximal coverage} conjunctions of size $k$,
i.e. conjunctions which entail subsets of the positive examples not included between each other. 

 \begin{algorithm}[t]
\footnotesize
{
\begin{myalgorithm}[]
def join($\mathcal{P}$, bestsol, $E^+$, $E^-$, k):
  if bestsol == None:
    return incomplete_join($\mathcal{P}$, $E^+$, $E^-$)
  else:
    return complete_join($\mathcal{P}$, $E^+$, $E^-$,k)

def incomplete_join($\mathcal{P}$, $E^+$, $E^-$):
  uncovered, conjunctions = $E^+$, {}
  while uncovered:
    encoding = buildencoding($\mathcal{P}$, $E^+$, $E^-$, conjunctions)
    conj = conj_max_coverage(uncovered, encoding)
    if not conj:
      break
    uncovered -= pos_entailed(conj)
    conjunctions += {conj}
  return conjunctions
    
def complete_join($\mathcal{P}$, $E^+$, $E^-$, k):
  conjunctions = {}
  while True:
    encoding = buildencoding($\mathcal{P}$, $E^+$, $E^-$, conjunctions)
    encoding += sizeconstraint(k)
    $\tau$ = find_assignment(encoding)
    if not $\tau$:
      break
    while True:
      assignment = cover_more_pos(encoding, $\tau$)
      if not assignment:
        break
      $\tau$ = assignment
    conjunctions += {conjunction($\tau$)}
  return conjunctions
\end{myalgorithm}
\caption{
Join stage
}
\label{alg:join}
}
\end{algorithm}

\begin{myexample}[\textbf{Join stage}]
Consider the positive examples $E^+=\{f([a,b,c,d]), f([c,b,d,e])\}$, the negative examples $E^-=\{f([c,b]), f([d,b]), f([a,c,d,e])\}$, and the programs:
\[
    \begin{array}{l}
    p_1 = \left\{
    \begin{array}{l}
\emph{f(S) $\leftarrow$ head(S,a)}
    \end{array}
    \right\}\\
    p_2 = \left\{
    \begin{array}{l}
\emph{f(S) $\leftarrow$ last(S,e)}
    \end{array}
    \right\}\\
    p_3 = \left\{
    \begin{array}{l}
\emph{f(S) $\leftarrow$ tail(S,T), head(T,b)}
    \end{array}
    \right\}\\
    p_4 = \left\{
    \begin{array}{l}
\emph{f(S) $\leftarrow$ head(S,c)}\\
\emph{f(S) $\leftarrow$ tail(S,T), f(T)}
    \end{array}
    \right\}\\
    p_5 = \left\{
    \begin{array}{l}
\emph{f(S) $\leftarrow$ head(S,d)}\\
\emph{f(S) $\leftarrow$ tail(S,T), f(T)}
    \end{array}
    \right\}
    \end{array}
\]
Each of these programs entails at least one positive and one negative example.
The incomplete join stage first outputs the conjunction $c_1=\{p_3, p_4, p_5\}$ as it entails all the positive examples and no negative example.
The complete join stage then outputs the conjunctions $c_2=\{p_1, p_3\}$ and $c_3=\{p_2, p_3\}$.
The other conjunctions are not output because they (i) do not entail any positive example, or (ii) entail some negative example, or (iii) are subsumed by $c_2$ or $c_3$.
% While $\{c_1\}$ has size 13, $\{c_2,c_3\}$ has size 10.
\end{myexample}
 
\subsubsection{Finding Conjunctions using SAT}
To find conjunctions, we use a Boolean satisfiability~(SAT)~\cite{sat-handbook} approach.
We build a propositional encoding (lines 10 and 21) for the join stage as follows.
Let $\mathcal{P}$ be the set of input programs.
For each program $h\in \mathcal{P}$, the variable $p_h$ indicates that $h$ is in a conjunction.
For each positive example $e \in E^+$, the variable $c_e$ indicates that the conjunction entails $e$.
The constraint
$F^+_e = c_e \rightarrow \bigwedge_{h \in \mathcal{P} \mid B \cup h \not\models e} \neg p_h$ ensures that if the conjunction entails $e$, then every program in the conjunction entails $e$.
For each negative example $e \in E^-$, the constraint
$F^-_e =\bigvee_{h \in \mathcal{P} \mid B \cup h \not\models e} p_h$ ensures that at least one of the programs in the conjunction does not entail $e$.

\paragraph{Subset-maximal coverage conjunctions.} 
For the complete join stage, to find all conjunctions of size $k$ with subset-maximal coverage,
we use a SAT solver to enumerate maximal satisfiable subsets~\cite{DBLP:journals/jar/LiffitonS08} corresponding to the subset-maximal coverage conjunctions on the following propositional encoding.
% on our propositional encoding.
% \ac{I do not understand this sentence ``enumerating subsets ... on our propositional encoding does not make sense''. Can we drop ''on our proposition encoding''?}
We build the constraint $S = \sum_h size(h) \cdot p_h \leq k$ to bound the size of the conjunctions and we encode $S$ as a propositional formula $F_S$~\cite{DBLP:conf/ki/MantheyPS14}.
To find a conjunction with subset-maximal coverage, we iteratively call a SAT solver on the formula $F = \bigwedge_{e \in E^+} F^+_e \wedge \bigwedge_{e \in E^-} F^-_e \land F_S$.
If $F$ has a satisfying assignment $\tau$ (line 23), we form a conjunction $c$ by including a program $h$ iff $\tau(h) = 1$
and update $F$ to $F \land \bigwedge_{e \in E^+ \mid B \cup c \models e} c_e \wedge \bigvee_{e \in E^+ \mid B \cup c \not\models e} c_e$ to ensure that subsequent conjunctions cover more examples (line 27).
We iterate until $F$ is unsatisfiable (lines 26 to 30), in which case $c$ has subset-maximal coverage.
To enumerate \emph{all} conjunctions, we iteratively call this procedure on the formula $F \land \bigwedge_{c \in C} \bigvee_{e \in E^+ \mid B \cup c \not\models e} c_e$, where $C$ is the set of conjunctions found so far.

\paragraph{Maximal coverage conjunctions.}
For the incomplete join stage, we use maximum satisfiability (MaxSAT)~\cite{maxsat} solving to find conjunctions which entail the maximum number of uncovered positive examples (line 11).
The MaxSAT encoding includes the hard clauses $\bigwedge_{e \in E^+} F^+_e \wedge \bigwedge_{e \in E^-} F^-_e$
to ensure correct coverage, as well as soft clauses $(c_e)$ for each uncovered positive example $e$
to maximise the number of uncovered examples.

\subsection{Constrain}
In the constrain stage (line 20 in Algorithm \ref{alg:metaspec}), \name{} uses two optimally sound constraints to prune the hypothesis space. 
If a hypothesis does not entail any positive example, \name{} prunes all its specialisations, as they cannot be in a conjunction in an optimal solution:

\begin{proposition}
% [\textbf{Specialisation constraint when tp=0}]
\label{prop:totincomplete}
Let $h_1$ be a hypothesis that does not entail any positive example
and $h_2$ be a specialisation of $h_1$.
Then $h_2$ cannot be in a conjunction in an optimal solution.
\end{proposition}
\noindent
If a hypothesis does not entail any negative example, \name{} prunes all its specialisations, as they cannot be in a conjunction in an optimal solution:
\begin{proposition}
% [\textbf{Specialisation constraint when fp=0}]
\label{prop:consistent}
Let $h_1$ be a hypothesis that does not entail any negative example and $h_2$ be a specialisation of $h_1$.
Then $h_2$ cannot be in a conjunction in an optimal solution.
\end{proposition}
\noindent
The appendix includes proofs for these propositions.

\subsection{Correctness}
We prove the correctness of \name{}:
\begin{theorem}
% [\textbf{Correctness}] 
\label{prop:correct}
\name{} returns an optimal solution, if one exists.
\end{theorem}
\noindent
The proof is in the appendix. 
To show this result, we show that (i) \name{} can generate and test every non-splittable program, (ii) each rule of an optimal solution is equivalent to the conjunction of non-splittable rules, and (iii) our constraints (Propositions \ref{prop:subsumed}, \ref{prop:totincomplete}, and \ref{prop:consistent}) never prune optimal solutions.
% \ch{need to clarify: we learn an optimal program in the hypothesis considered (there might exists a shorter \emph{splittable} program that we cannot find)}

\section{Experiments}
To test our claim that our join stage can improve learning performance, our experiments aim to answer the question:
\begin{enumerate}
\item[\textbf{Q1}] Can the join stage improve learning performance?
\end{enumerate}
To answer \textbf{Q1}, we compare learning with and without the join stage.

To test our claim that eliminating splittable programs in the generate stage can improve learning performance, our experiments aim to answer the question:
\begin{enumerate}
\item[\textbf{Q2}] Can eliminating splittable programs in the generate stage improve learning performance?
\end{enumerate}
To answer \textbf{Q2}, we compare learning with and without the constraint eliminating splittable programs.

To test our claim that \name{} can learn programs with big splittable rules, our experiments aim to answer the question:
\begin{enumerate}
\item[\textbf{Q3}] How well does \name{} scale with the size of splittable rules?
\end{enumerate}
To answer \textbf{Q3}, we vary the size of rules and evaluate the performance of \name{}.

Finally, to test our claim that \name{} can outperform other approaches, our experiments aim to answer the question:

\begin{enumerate}
\item[\textbf{Q4}] How well does \name{} compare against other approaches?
\end{enumerate}
To answer \textbf{Q4}, we compare \name{} against \combo{} \cite{combo} and \ale{} \cite{aleph} on multiple tasks and domains\footnote{
We also tried rule selection approaches (Section \ref{sec:related}) but precomputing every possible rule is infeasible on our datasets. 
}.

\subsubsection{Domains}
We consider several domains.
The appendix provides additional information about our domains and tasks.

\noindent
\textbf{IGGP.} In inductive general game playing (IGGP) \cite{iggp}, the task is to learn rules from game traces from the general game playing competition \cite{ggp}.

\noindent
 \textbf{Zendo.}
 Zendo is an inductive game where the goal is to identify a secret rule that structures must follow \cite{zendo,combo}.

\noindent
 \textbf{IMDB.} The international movie database (IMDB) \cite{mihalkova2007} is a relational domain containing relations between movies, directors, and actors.

\noindent
 \textbf{Pharmacophores.} 
 The task is to identify structures responsible for the medicinal activity of molecules \cite{finn1998}.

\noindent
 \textbf{Strings.} The goal is to identify recursive patterns to classify strings. 

\noindent
 \textbf{1D-ARC.} This dataset \cite{onedarc} contains visual reasoning tasks inspired by the abstract reasoning corpus \cite{arc}.

% \begin{figure}
% \small
% \begin{tabular}{l}
% \emph{f(A) $\leftarrow$  f$_{1}$(A), f$_{2}$(A)}\\
% \emph{f$_{1}$(A) $\leftarrow$  head(A,B), a(B)}\\
% \emph{f$_{1}$(A) $\leftarrow$  tail(A,B), f$_{1}$(B)}\\
% \emph{f$_{2}$(A) $\leftarrow$  head(A,B), b(B)}\\
% \emph{f$_{2}$(A) $\leftarrow$  head(A,B), f$_{2}$(B)}\\
% \end{tabular}
% % \caption{Example of hypothesis for the \emph{string} domain. This hypothesis states that a string is positive if it contains both the letter \emph{a} followed by the letter \emph{b} and the letter \emph{c} followed by the letter \emph{d}. We vary the number of letters to identify in a string hence the number of body literals.}
% \label{hypstring}
% \end{figure}

\subsubsection{Experimental Setup}
We use 60s and 600s timeouts.
We repeat each experiment 5 times. 
We measure predictive accuracies (the proportion of correct predictions on testing data). 
For clarity, our figures only show tasks where the approaches differ. 
The appendix contains the detailed results for each task.
We use an 8-core 3.2 GHz Apple M1 and a single CPU to run the experiments. 
We use the MaxSAT solver UWrMaxSat \cite{uwrmaxsat} and the SAT solver CaDiCaL 1.5.3 \cite{cadical} (via PySAT~\cite{pysat}) in the join stage of \name{}.

\textbf{Q1.}
We compare learning with and without the join stage.
To isolate the impact of the join stage, we allow splittable programs in the generate stage. 
% In other words, the experimental variable is whether we use the join stage.

\textbf{Q2.}
We compare the predictive accuracies of \name{} with and without the constraint that eliminates splittable programs.

\textbf{Q3.}
To evaluate scalability, for increasing values of $k$, we generate a task where an optimal solution has size $k$. We learn a hypothesis with a single splittable rule. 
We use a \textit{zendo} task similar to the one shown in Section \ref{introduction} and a \textit{string} task.

\textbf{Q4.}
We provide \name{} and \combo{} with identical input. 
The only differences are (i) the join stage, and (ii) the elimination of splittable programs in the generate stage. 
However, because it can build conjunctions in the join stage, \name{} searches a larger hypothesis space than \combo{}.
\ale{} uses a different bias than \name{} to define the hypothesis space. 
In particular, \ale{} expects a maximum rule size as input.
Therefore, the comparison is less fair and should be viewed as indicative only. 

%\begin{figure}[ht]
%\centering
%\small
%\begin{tabular}{l}
%    \emph{zendo(S) $\leftarrow$ piece(S,R), red(R),upright(R),piece(S,B),blue(B),}\\
%  \hspace{44pt} \emph{reversed(B),piece(S,P), pink(P),side(P)}
%    \end{tabular}
%\caption{Example \emph{zendo} hypothesis. 
%A positive Zendo structure contains an upright red piece, a reversed blue one, and a pink one on its side. 
%We vary the number of pieces to identify, hence the size of optimal solutions.
%}
%\label{hypzendo}
%\end{figure}
\paragraph{Reproducibility.}
The code and experimental data for reproducing the experiments are provided as supplementary material and will be made publicly available if the paper is accepted for publication.

\definecolor{mygreen1}{cmyk}{0.9,0,1.0,0.05}
\definecolor{mygreen}{rgb}{0.3,0.5,0.8}
\definecolor{mygray}{rgb}{0.6,0.6,0.6}

% \subsection{Experiment 1: Join Stage}
\subsection*{Experimental Results}

\subsubsection{Q1. Can the Join Stage Improve Performance?}
\begin{figure}[!ht]
  \begin{minipage}{0.235\textwidth}
\begin{tikzpicture}
\begin{axis}[%
  xmin=45,
  xmax=105,
  ymin=45,
  ymax=105,
  width=\textwidth,
  height=\textwidth,
xtick={50, 60, 70, 80, 90, 100},
ytick={50, 60, 70, 80, 90, 100},
tick label style={font=\scriptsize},
grid=both,
xlabel={without join stage},
ylabel={with join stage},
xlabel style={font=\scriptsize},
ylabel style={font=\scriptsize},
xlabel near ticks,
ylabel near ticks,
scatter/classes={%
    attrition={mark=o}}]
\addplot[scatter,
    only marks,%
    mygreen,
    fill opacity=0.4,
   % draw opacity=0,
    scatter src=explicit symbolic,
    visualization depends on = {\thisrow{count} \as \perpointmarksize},
    scatter/@pre marker code/.append style={/tikz/mark size=\perpointmarksize}]%
table[meta=label] {
x y label count 
66 75 label 1.0
67 75 label 1.4142135623730951
91 100 label 1.7320508075688772
100 90 label 1.7320508075688772
62 100 label 1.7320508075688772
67 61 label 1.7320508075688772
50 100 label 3.872983346207417
50 82 label 1.4142135623730951
50 81 label 1.0
50 62 label 1.0
50 52 label 2.0
50 58 label 1.0
50 96 label 1.7320508075688772
50 98 label 1.0
50 97 label 1.4142135623730951
92 100 label 1.0
95 99 label 1.0
71 94 label 1.0
70 95 label 1.0
68 93 label 1.0
83 81 label 1.0
75 79 label 1.0
78 80 label 1.0
90 96 label 1.0
90 98 label 1.0
96 100 label 1.0
50 53 label 1.7320508075688772
50 54 label 1.0
50 55 label 1.7320508075688772
50 51 label 1.0
50 56 label 1.4142135623730951
50 46 label 1.0
58 100 label 1.4142135623730951
98 100 label 1.0
60 100 label 1.0
50 94 label 1.0
50 90 label 1.0
58 75 label 1.0
52 100 label 1.0
50 83 label 1.0
62 98 label 1.0
52 96 label 1.0
53 97 label 1.0
50 99 label 1.0
52 56 label 1.0
53 62 label 1.0
50 64 label 1.0
50 75 label 1.0
50 92 label 1.0
83 100 label 1.4142135623730951
54 72 label 1.0
50 57 label 1.0
53 100 label 1.0
52 97 label 1.0
52 88 label 1.0
    };
\draw [mygray,dashed] (rel axis cs:0,0) -- (rel axis cs:1,1);
\end{axis}
\end{tikzpicture}
\end{minipage}
  \begin{minipage}{0.235\textwidth}
\begin{tikzpicture}
\begin{axis}[%
  xmin=45,
  xmax=105,
  ymin=45,
  ymax=105,
  width=\textwidth,
  height=\textwidth,
%  scale only axis,
xtick={50, 60, 70, 80, 90, 100},
ytick={50, 60, 70, 80, 90, 100},
tick label style={font=\scriptsize},
grid=both,
xlabel={without join stage},
ylabel={with join stage},
xlabel style={font=\scriptsize},
ylabel style={font=\scriptsize},
xlabel near ticks,
ylabel near ticks,
scatter/classes={%
    attrition={mark=o}}]
\addplot[scatter,
    only marks,%
    mygreen,
    fill opacity=0.4,
   % draw opacity=0,
    scatter src=explicit symbolic,
    visualization depends on = {\thisrow{count} \as \perpointmarksize},
    scatter/@pre marker code/.append style={/tikz/mark size=\perpointmarksize}]%
table[meta=label] {
x y label count 
67 75 label 1.7320508075688772
62 100 label 1.0
67 100 label 1.7320508075688772
52 100 label 1.7320508075688772
77 100 label 1.7320508075688772
50 100 label 3.7416573867739413
50 98 label 1.4142135623730951
50 94 label 1.0
50 97 label 1.4142135623730951
76 97 label 1.0
72 94 label 1.0
74 90 label 1.0
70 94 label 1.0
78 73 label 1.0
79 81 label 1.0
83 80 label 1.0
94 97 label 1.0
91 99 label 1.0
92 100 label 1.0
50 58 label 1.4142135623730951
50 55 label 1.4142135623730951
51 57 label 1.0
50 51 label 1.0
48 100 label 1.4142135623730951
58 100 label 1.4142135623730951
98 100 label 1.0
60 100 label 1.0
75 100 label 1.0
90 100 label 1.0
50 96 label 1.0
50 75 label 1.4142135623730951
58 91 label 1.0
60 99 label 1.0
50 83 label 1.0
50 52 label 1.4142135623730951
62 98 label 1.0
52 96 label 1.0
53 97 label 1.0
50 99 label 1.0
50 53 label 1.4142135623730951
52 56 label 1.0
53 62 label 1.0
50 64 label 1.4142135623730951
50 92 label 1.0
50 89 label 1.0
83 100 label 1.4142135623730951
54 73 label 1.0
50 77 label 1.0
50 78 label 1.0
50 65 label 1.0
50 85 label 1.0
50 57 label 1.0
53 100 label 1.0
52 97 label 1.0
52 88 label 1.0
    };
\draw [mygray,dashed] (rel axis cs:0,0) -- (rel axis cs:1,1);
\end{axis}
\end{tikzpicture}
\end{minipage}
\caption{Predictive accuracy (\%) with and without join stage with 60s (left) and 600s (right) timeouts.}
\label{fig:nojoin}
\end{figure}
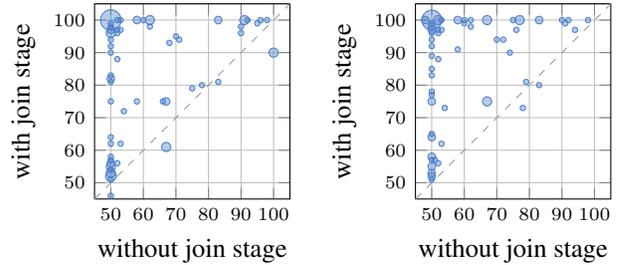

Figure \ref{fig:nojoin} shows that the join stage can drastically improve predictive accuracies. 
A McNeymar's test confirms ($p<0.01$) that the join stage improves accuracies on 24/42 tasks with a 60s timeout and on 22/42 tasks with a 600s timeout. 
There is no significant difference for the other tasks.

The join stage can learn big rules which otherwise cannot be learned.
For instance, for the task \emph{pharma1}, the join stage finds a rule of size 17 which has 100\% accuracy. 
By contrast, without the join stage, no solution is found, resulting in default accuracy (50\%).
Similarly, an optimal solution for the task \emph{iggp-rainbow} has a single rule of size 19. 
This rule is splittable and is the conjunction of 6 small rules.
The join stage identifies this rule in less than 1s as it entails all the positive examples.
By contrast, without the join stage, the system exceeds the timeout without finding a solution as it needs to search through the set of all rules up to size 19 to find a solution.

The overhead of the join stage is small. 
For instance, for the task \emph{scale} in the \emph{1D-ARC} domain, the join stage takes less than 1\% of the total learning time, yet this stage allows us to find a perfectly accurate solution with a rule of size 13.

% Overall, the results suggest that the join stage can drastically improve predictive accuracy as the answer to \textbf{Q1}.
Overall, the results suggest that the answer to \textbf{Q1} is that the join stage can substantially improve predictive accuracy.

\subsubsection{Q2. Can Eliminating Splittable Programs Improve Performance?}

%\begin{figure}
%\begin{minipage}{0.2\textwidth}
%\begin{tikzpicture}
%\begin{axis}[
%boxplot/draw direction=y,
%ylabel={Accuracy},
%xlabel={Timeout (s)},
%xlabel style={font=\small},
%ylabel style={font=\small},
%height=4cm,
% ymin=0,ymax=7,
%cycle list={{red},{blue}},
%boxplot={
%        draw position={1/3 + floor(\plotnumofactualtype/2) + 1/3*mod(\plotnumofactualtype,2)},
%        box extend=0.3
%},
%x=1.25cm,
%xtick={1/2, 3/2, 5/2, 7/2, 9/2},
%ytick={50, 60, 70, 80, 90, 100},
% x tick label as interval,
%ymin=45,
%ymax=105,
%xmin=0,
%xmax=5,
%xticklabels={30, 60, 90, 120, 150, 180},
%yticklabels={50, 60, 70, 80, 90, 100},
%x tick label style={
%        text width=2.5cm,
%        align=center
%},
%]
%\addplot+ [boxplot] table [y index=0] {\mydatatimeout};
%\addplot+ [boxplot] table [y index=1] {\mydatatimeout};
%\addplot+ [boxplot] table [y index=2] {\mydatatimeout};
%\addplot+ [boxplot] table [y index=3] {\mydatatimeout};
%\addplot+ [boxplot] table [y index=4] {\mydatatimeout};
%\addplot+ [boxplot] table [y index=5] {\mydatatimeout};
%\addplot+ [boxplot] table [y index=1] {\mydatatimeout};
%\addplot+ [boxplot] table [y index=1] {\mydatatimeout};
%\addplot+ [boxplot] table [y index=1] {\mydatatimeout};
%\addplot+ [boxplot] table [y index=1] {\mydatatimeout};
% \addplot table [y index=1] {\mydatatimeout}
%\end{axis}
%\end{tikzpicture}
%\end{minipage}
%\caption{Accuracy for increasingly large timeout.
%}
%\label{fig:nononsplittable}
%\end{figure}

\begin{figure}[!ht]
  \begin{minipage}{0.235\textwidth}
\begin{tikzpicture}
\begin{axis}[%
  xmin=45,
  xmax=105,
  ymin=45,
  ymax=105,
  width=\textwidth,
  height=\textwidth,
xtick={50, 60, 70, 80, 90, 100},
ytick={50, 60, 70, 80, 90, 100},
tick label style={font=\scriptsize},
grid=both,
xlabel={with splittable},
ylabel={without splittable},
xlabel style={font=\scriptsize},
ylabel style={font=\scriptsize},
xlabel near ticks,
ylabel near ticks,
scatter/classes={%
    attrition={mark=o}}]
\addplot[scatter,
    only marks,%
    mygreen,
    fill opacity=0.4,
   % draw opacity=0,
    scatter src=explicit symbolic,
    visualization depends on = {\thisrow{count} \as \perpointmarksize},
    scatter/@pre marker code/.append style={/tikz/mark size=\perpointmarksize}]%
table[meta=label] {
x y label count 
75 67 label 1.7320508075688772
90 100 label 1.4142135623730951
100 83 label 1.0
100 81 label 1.0
100 80 label 1.0
61 100 label 1.7320508075688772
82 100 label 1.4142135623730951
81 100 label 1.0
50 99 label 1.7320508075688772
62 57 label 1.0
52 57 label 1.0
58 56 label 1.0
100 99 label 1.0
96 95 label 1.0
99 96 label 1.0
95 93 label 1.0
93 95 label 1.0
80 81 label 1.0
46 100 label 1.0
50 100 label 1.0
56 100 label 1.0
52 98 label 1.0
96 98 label 1.0
100 89 label 1.0
72 73 label 1.0
    };
\draw [mygray,dashed] (rel axis cs:0,0) -- (rel axis cs:1,1);
\end{axis}
\end{tikzpicture}
% \caption{Accuracy with and without generating non-splittable programs with a 60s timeout.}
\label{fig:nononsplittable60}
\end{minipage}
  \begin{minipage}{0.235\textwidth}
\begin{tikzpicture}
\begin{axis}[%
  xmin=45,
  xmax=105,
  ymin=45,
  ymax=105,
  width=\textwidth,
  height=\textwidth,
%  scale only axis,
xtick={50, 60, 70, 80, 90, 100},
ytick={50, 60, 70, 80, 90, 100},
tick label style={font=\scriptsize},
grid=both,
xlabel={with splittable},
ylabel={without splittable},
xlabel style={font=\scriptsize},
ylabel style={font=\scriptsize},
xlabel near ticks,
ylabel near ticks,
scatter/classes={%
    attrition={mark=o}}]
\addplot[scatter,
    only marks,%
    mygreen,
    fill opacity=0.4,
   % draw opacity=0,
    scatter src=explicit symbolic,
    visualization depends on = {\thisrow{count} \as \perpointmarksize},
    scatter/@pre marker code/.append style={/tikz/mark size=\perpointmarksize}]%
table[meta=label] {
x y label count 
75 67 label 1.7320508075688772
100 90 label 1.0
98 99 label 1.0
98 96 label 1.0
94 96 label 1.0
100 96 label 1.0
100 99 label 1.0
94 100 label 1.4142135623730951
90 100 label 1.0
73 75 label 1.0
80 81 label 1.0
58 100 label 1.4142135623730951
55 100 label 1.4142135623730951
57 100 label 1.0
51 100 label 1.0
96 100 label 1.4142135623730951
100 75 label 1.0
75 100 label 1.0
55 75 label 1.0
91 74 label 1.0
99 60 label 1.0
83 100 label 1.0
52 56 label 1.0
97 100 label 1.4142135623730951
98 100 label 1.0
99 83 label 1.0
52 94 label 1.0
53 76 label 1.0
62 74 label 1.0
56 78 label 1.0
62 93 label 1.0
60 93 label 1.0
64 86 label 1.0
75 86 label 1.0
92 100 label 1.0
89 94 label 1.0
100 83 label 1.0
53 67 label 1.0
73 83 label 1.0
77 76 label 1.0
50 69 label 1.0
78 92 label 1.0
65 53 label 1.0
57 87 label 1.0
97 98 label 1.0
88 100 label 1.0
    };
\draw [mygray,dashed] (rel axis cs:0,0) -- (rel axis cs:1,1);
\end{axis}
\end{tikzpicture}
\label{fig:nononsplittable600}
\end{minipage}
\caption{Predictive accuracies (\%) with and without generating splittable programs with 60s (left) and 600s (right) timeouts.
}
\label{fig:nononsplittable}
\end{figure}
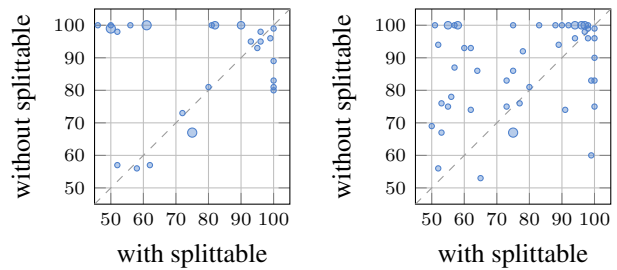

Figure \ref{fig:nononsplittable} shows that eliminating splittable programs in the generate stage can improve learning performance. 
A McNeymar's test confirms ($p<0.01$) that eliminating splittable programs improves performance on 8/42 tasks with a 60s timeout and on 11/42 tasks with a 600s timeout.
It degrades performance ($p<0.01$) on 1/42 tasks with a 60s timeout.
There is no significant difference for the other tasks.

Eliminating splittable programs from the generate stage can greatly reduce the number of programs \name{} considers. 
For instance, for \emph{iggp-rainbow}, the hypothesis space contains 1,986,422 rules of size at most 6, but only 212,564 are non-splittable.
Likewise, for \emph{string2}, when eliminating splittable programs, \name{} finds a perfectly accurate solution in 133s (2min13s).
By contrast, with splittable programs, \name{} considers more programs and fails to find a solution within the 600s timeout, resulting in default accuracy (50\%).

Overall, these results suggest that the answer to \textbf{Q2} is that eliminating splittable programs from the generate stage can improve predictive accuracies. 

\subsubsection{Q3. How Well Does \name{} Scale?}

\pgfplotstableread[col sep=comma]{./figures/zendo/zendo_body_size_2_aleph_parameter.csv}\zendoaleph
\pgfplotstableread[col sep=comma]{./figures/zendo/zendo_body_size_2_metaspecial_parameter.csv}\zendometaspec
\pgfplotstableread[col sep=comma]{./figures/zendo/zendo_body_size_2_combo_parameter.csv}\zendocombo

\pgfplotstableread[col sep=comma]{./figures/strings/stringbodysize_1_aleph_parameter.csv}\stringaleph
\pgfplotstableread[col sep=comma]{./figures/strings/stringbodysize_1_metaspecial_parameter.csv}\stringmetaspec
\pgfplotstableread[col sep=comma]{./figures/strings/stringbodysize_1_combo_parameter.csv}\stringcombo

\begin{figure}[!ht]
  \begin{minipage}{0.05\textwidth}
\begin{tikzpicture}
\begin{customlegend}[legend columns=5,legend style={nodes={scale=1, transform shape},align=left,column sep=0ex},
        legend entries={\name, \combo, \ale}]
        \addlegendimage{red,mark=diamond*}
        \addlegendimage{blue,mark=square*}
        \addlegendimage{mygreen1,mark=triangle*}
\end{customlegend}
\end{tikzpicture}
\end{minipage}\hfill\\
% \begin{figure}
  \begin{minipage}{0.23\textwidth}
\begin{tikzpicture}[scale=0.42]
\begin{axis}[
  legend style={at={(0.5,0.35)},anchor=west},
   legend style={font=\large},
  tick label style={font=\huge},
  xlabel style={font=\huge},
  ylabel style={font=\huge},
  xlabel=Optimal solution size,
  ylabel=Accuracy,
  xmin=4,
  xmax=103,
  ymin=48,
  ymax=102,
  %  log ticks with fixed point,
  ]
\addplot[mark=diamond*,red,
                error bars/.cd,
                y dir=both,
                error mark,
                y explicit]table[x=xs,y=acc_av,y error=acc_std] {\zendometaspec};
\addplot[blue,mark=square*,
                error bars/.cd,
                y dir=both,
                error mark,
                y explicit]table[x=xs,y=acc_av,y error=acc_std] {\zendocombo};
\addplot[mygreen1,mark=triangle*,
                error bars/.cd,
                y dir=both,
                error mark,
                y explicit]table[x=xs,y=acc_av,y error=acc_std] {\zendoaleph};
\end{axis}
\end{tikzpicture}
% \caption{Accuracy versus the number of body literals in target hypotheses for \emph{zendo}.}
\label{fig:acczendo}
\end{minipage}
\hfill
  \begin{minipage}{0.23\textwidth}
\begin{tikzpicture}[scale=0.42]
\begin{axis}[
  legend style={at={(0.5,0.35)},anchor=west},
   legend style={font=\large},
  tick label style={font=\huge},
  xlabel style={font=\huge},
  ylabel style={font=\huge},
  xlabel=Optimal solution size,
  ylabel=Accuracy,
  xmin=6,
  xmax=100,
  ymin=48,
  ymax=102,
  %  log ticks with fixed point,
  ]
\addplot[red,mark=diamond*,
                error bars/.cd,
                y dir=both,
                error mark,
                y explicit]table[x=xs,y=acc_av,y error=acc_std] {\stringmetaspec};
\addplot[blue,mark=square*,
                error bars/.cd,
                y dir=both,
                error mark,
                y explicit]table[x=xs,y=acc_av,y error=acc_std] {\stringcombo};
\addplot[mygreen1,mark=triangle*,
                error bars/.cd,
                y dir=both,
                error mark,
                y explicit]table[x=xs,y=acc_av,y error=acc_std] {\stringaleph};

\end{axis}
\end{tikzpicture}
\label{fig:accstrings}
\end{minipage}

\caption{Predictive accuracies (\%) when varying the optimal solution size for \emph{zendo} (left) and \emph{string} (right) with a 600s timeout.
}
\label{fig:bodylits}
\end{figure}
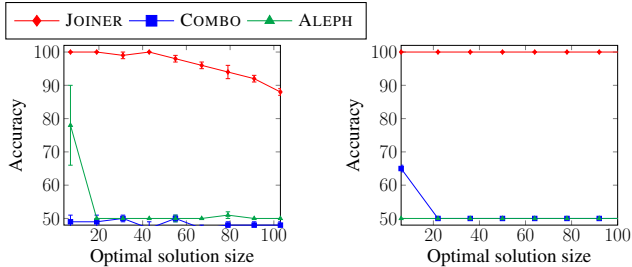

Figure \ref{fig:bodylits} shows that \name{} can learn an almost perfectly accurate hypothesis with up to 100 literals for both the \emph{zendo} and \emph{string} tasks. 
By contrast, \combo{} and \ale{} struggle to learn hypotheses with more than 10 literals.

\name{} learns a \emph{zendo} hypothesis of size $k$ after searching for programs of size 4.
By contrast, \combo{} must search for programs up to size $k$ to find a solution. 
Similarly, an optimal solution for the \emph{string} task is the conjunction of programs with 6 literals each. 
By contrast, \combo{} must search for programs up to size $k$ to find a solution. 
\ale{} struggles to learn recursive programs and thus struggles on the \emph{string} task. 

Overall, the results suggest that the answer to \textbf{Q3} is that \name{} can scale well with the size of rules.
% , with up to 100 literals.

\subsubsection{Q4. How Does \name{} Compare to Other Approaches?}

Table \ref{tab:agg_acc} shows the predictive
accuracies aggregated over each domain. 
It shows \name{} achieves higher accuracies than \combo{} and \ale{} on almost all domains. 

\begin{table}[!ht]
\small
\centering
\begin{tabular}{@{}l|ccc@{}}
\textbf{Task} & \textbf{\ale{}} & \textbf{\combo{}} & \textbf{\name{}}\\
\midrule
\emph{iggp} & 78 $\pm$ 3 & 86 $\pm$ 2 & \textbf{96 $\pm$ 1}\\
\emph{zendo} & \textbf{100 $\pm$ 0} & 86 $\pm$ 3 & 94 $\pm$ 2\\
\emph{pharma} & 50 $\pm$ 0 & 53 $\pm$ 2 & \textbf{98 $\pm$ 1}\\
\emph{imdb} & 67 $\pm$ 6 & \textbf{100 $\pm$ 0} & \textbf{100 $\pm$ 0}\\
\emph{string} & 50 $\pm$ 0 & 50 $\pm$ 0 & \textbf{100 $\pm$ 0}\\
\emph{onedarc} & 51 $\pm$ 1 & 57 $\pm$ 2 & \textbf{89 $\pm$ 1}\\
\end{tabular}
\caption{
Aggregated predictive accuracies (\%) with a 600s timeout.
}
\label{tab:agg_acc}
\end{table}

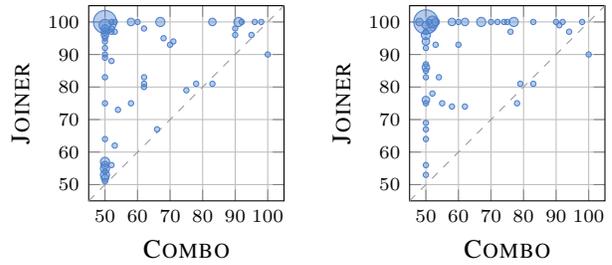
\begin{figure}[!ht]
  \begin{minipage}{0.235\textwidth}
\begin{tikzpicture}
\begin{axis}[%
  xmin=45,
  xmax=105,
  ymin=45,
  ymax=105,
  width=\textwidth,
  height=\textwidth,
xtick={50, 60, 70, 80, 90, 100},
ytick={50, 60, 70, 80, 90, 100},
tick label style={font=\scriptsize},
grid=both,
xlabel={\combo{}},
ylabel={\name{}},
xlabel style={font=\scriptsize},
ylabel style={font=\scriptsize, yshift=-2mm},
xlabel near ticks,
ylabel near ticks,
scatter/classes={%
    attrition={mark=o}}]
\addplot[scatter,
    only marks,%
    mygreen,
    fill opacity=0.4,
   % draw opacity=0,
    scatter src=explicit symbolic,
    visualization depends on = {\thisrow{count} \as \perpointmarksize},
    scatter/@pre marker code/.append style={/tikz/mark size=\perpointmarksize}]%
table[meta=label] {
x y label count 
66 67 label 1.0
91 100 label 1.7320508075688772
100 90 label 1.0
62 83 label 1.0
62 81 label 1.0
62 80 label 1.0
67 100 label 1.7320508075688772
50 100 label 4.358898943540674
50 99 label 2.23606797749979
50 57 label 1.7320508075688772
50 56 label 1.4142135623730951
50 96 label 1.4142135623730951
50 95 label 1.0
50 98 label 1.4142135623730951
50 97 label 1.4142135623730951
92 100 label 1.0
95 96 label 1.0
71 94 label 1.0
70 93 label 1.0
68 95 label 1.0
83 81 label 1.0
75 79 label 1.0
78 81 label 1.0
90 96 label 1.0
90 98 label 1.0
96 100 label 1.0
50 53 label 1.7320508075688772
50 54 label 1.0
50 55 label 1.7320508075688772
50 51 label 1.0
50 52 label 1.4142135623730951
58 100 label 1.4142135623730951
98 100 label 1.0
60 100 label 1.0
50 94 label 1.0
50 90 label 1.0
58 75 label 1.0
52 100 label 1.0
50 83 label 1.0
62 98 label 1.0
52 98 label 1.0
53 97 label 1.0
52 56 label 1.0
53 62 label 1.0
50 64 label 1.0
50 75 label 1.0
50 92 label 1.0
50 89 label 1.0
83 100 label 1.4142135623730951
54 73 label 1.0
53 100 label 1.0
52 97 label 1.0
52 88 label 1.0
    };
\draw [mygray,dashed] (rel axis cs:0,0) -- (rel axis cs:1,1);
\end{axis}
\end{tikzpicture}
\end{minipage}
  \begin{minipage}{0.235\textwidth}
\begin{tikzpicture}
\begin{axis}[%
  xmin=45,
  xmax=105,
  ymin=45,
  ymax=105,
  width=\textwidth,
  height=\textwidth,
xtick={50, 60, 70, 80, 90, 100},
ytick={50, 60, 70, 80, 90, 100},
tick label style={font=\scriptsize},
grid=both,
xlabel={\combo{}},
ylabel={\name{}},
xlabel style={font=\scriptsize},
ylabel style={font=\scriptsize, yshift=-2mm},
xlabel near ticks,
ylabel near ticks,
scatter/classes={%
    attrition={mark=o}}]
\addplot[scatter,
    only marks,%
    mygreen,
    fill opacity=0.4,
   % draw opacity=0,
    scatter src=explicit symbolic,
    visualization depends on = {\thisrow{count} \as \perpointmarksize},
    scatter/@pre marker code/.append style={/tikz/mark size=\perpointmarksize}]%
table[meta=label] {
x y label count 
100 90 label 1.0
62 100 label 1.4142135623730951
67 100 label 1.7320508075688772
52 100 label 2.23606797749979
77 100 label 1.7320508075688772
50 100 label 4.58257569495584
50 99 label 1.4142135623730951
50 96 label 1.7320508075688772
50 97 label 1.0
76 97 label 1.0
72 100 label 1.0
74 100 label 1.0
70 100 label 1.0
78 75 label 1.0
79 81 label 1.0
83 81 label 1.0
94 97 label 1.0
91 99 label 1.0
92 100 label 1.0
51 100 label 1.0
48 100 label 1.4142135623730951
58 100 label 1.4142135623730951
98 100 label 1.0
60 100 label 1.0
75 100 label 1.0
90 100 label 1.0
50 75 label 1.0
55 75 label 1.0
58 74 label 1.0
50 56 label 1.0
53 100 label 1.4142135623730951
50 83 label 1.0
50 94 label 1.4142135623730951
50 76 label 1.4142135623730951
62 74 label 1.0
52 78 label 1.0
53 93 label 1.0
60 93 label 1.0
50 86 label 1.4142135623730951
83 100 label 1.0
50 67 label 1.0
50 64 label 1.0
54 83 label 1.0
50 69 label 1.0
50 92 label 1.0
50 53 label 1.0
50 85 label 1.0
50 87 label 1.0
52 98 label 1.0
    };
\draw [mygray,dashed] (rel axis cs:0,0) -- (rel axis cs:1,1);
\end{axis}
\end{tikzpicture}
\end{minipage}
\caption{Predictive accuracies (\%) of \name{} versus \combo{} with 60s (left) and 600s (right) timeouts.}
\label{fig:combo}
\end{figure}

Figure \ref{fig:combo} shows that \name{} outperforms \combo{}.
A McNeymar's test confirms ($p<0.01$) that \name{} outperforms \combo{} on 27/42 tasks with a 60s timeout and on 26/42 tasks with a 600s timeout. 
\name{} and \combo{} have similar performance on other tasks.

\name{} can find hypotheses with big rules. 
For example, the \emph{flip} task in the \emph{1D-ARC} domain involves reversing the order of colored pixels in an image. 
\name{} finds a solution with two splittable rules of size 9 and 16.
By contrast, \combo{} only searches programs of size at most 4 before it timeouts and it does not learn any program.
\name{} can also perform better when learning non-splittable programs.
For instance, \name{} learns a perfectly accurate solution for \emph{iggp-rps} and proves that this solution is optimal in 20s. 
By contrast, \combo{} requires 440s (7min20s) to find the same solution and prove optimality.

\begin{figure}[!ht]
  \begin{minipage}{0.235\textwidth}
\begin{tikzpicture}
\begin{axis}[%
  xmin=44,
  xmax=106,
  ymin=44,
  ymax=106,
  width=\textwidth,
  height=\textwidth,
xtick={50, 60, 70, 80, 90, 100},
ytick={50, 60, 70, 80, 90, 100},
tick label style={font=\scriptsize},
grid=both,
xlabel={\ale{}},
ylabel={\name{}},
xlabel style={font=\scriptsize},
ylabel style={font=\scriptsize, yshift=-2mm},
xlabel near ticks,
ylabel near ticks,
scatter/classes={%
    attrition={mark=o}}]
\addplot[scatter,
    only marks,%
    mygreen,
    fill opacity=0.4,
   % draw opacity=0,
    scatter src=explicit symbolic,
    visualization depends on = {\thisrow{count} \as \perpointmarksize},
    scatter/@pre marker code/.append style={/tikz/mark size=\perpointmarksize}]%
table[meta=label] {
x y label count 
50 67 label 1.7320508075688772
50 100 label 6.557438524302
50 90 label 1.4142135623730951
50 83 label 1.4142135623730951
50 81 label 1.0
50 80 label 1.0
100 97 label 1.7320508075688772
97 100 label 1.7320508075688772
50 99 label 2.23606797749979
50 57 label 1.7320508075688772
50 56 label 1.7320508075688772
50 96 label 1.4142135623730951
50 95 label 1.0
50 98 label 2.0
50 97 label 2.0
100 96 label 1.0
100 94 label 1.0
100 93 label 1.0
100 95 label 1.0
99 81 label 1.4142135623730951
98 79 label 1.0
99 96 label 1.0
97 98 label 1.0
50 53 label 1.7320508075688772
50 54 label 1.0
50 55 label 2.0
50 51 label 1.0
50 52 label 1.4142135623730951
50 94 label 1.0
50 75 label 1.4142135623730951
50 60 label 1.4142135623730951
50 62 label 1.4142135623730951
50 64 label 1.0
50 92 label 1.0
50 89 label 1.0
50 73 label 1.0
50 88 label 1.0
    };
\draw [mygray,dashed] (rel axis cs:0,0) -- (rel axis cs:1,1);
\end{axis}
\end{tikzpicture}
\end{minipage}
  \begin{minipage}{0.235\textwidth}
\begin{tikzpicture}
\begin{axis}[%
  xmin=44,
  xmax=106,
  ymin=44,
  ymax=106,
  width=\textwidth,
  height=\textwidth,
xtick={50, 60, 70, 80, 90, 100},
ytick={50, 60, 70, 80, 90, 100},
tick label style={font=\scriptsize},
grid=both,
xlabel={\ale{}},
ylabel={\name{}},
xlabel style={font=\scriptsize},
ylabel style={font=\scriptsize, yshift=-2mm},
xlabel near ticks,
ylabel near ticks,
scatter/classes={%
    attrition={mark=o}}]
\addplot[scatter,
    only marks,%
    mygreen,
    fill opacity=0.4,
   % draw opacity=0,
    scatter src=explicit symbolic,
    visualization depends on = {\thisrow{count} \as \perpointmarksize},
    scatter/@pre marker code/.append style={/tikz/mark size=\perpointmarksize}]%
table[meta=label] {
x y label count 
50 67 label 2.0
100 90 label 1.0
50 100 label 7.3484692283495345
100 97 label 1.7320508075688772
65 100 label 1.7320508075688772
97 100 label 1.7320508075688772
50 99 label 1.4142135623730951
50 96 label 1.7320508075688772
50 97 label 1.4142135623730951
100 75 label 1.0
100 81 label 1.0
99 81 label 1.0
99 97 label 1.0
50 75 label 1.4142135623730951
92 74 label 1.0
59 60 label 1.0
50 56 label 1.0
50 83 label 1.7320508075688772
50 94 label 1.4142135623730951
50 76 label 1.4142135623730951
50 74 label 1.0
50 78 label 1.0
50 93 label 1.4142135623730951
50 86 label 1.4142135623730951
50 64 label 1.0
50 69 label 1.0
78 92 label 1.0
50 53 label 1.0
50 85 label 1.0
50 87 label 1.0
50 98 label 1.0
    };
\draw [mygray,dashed] (rel axis cs:0,0) -- (rel axis cs:1,1);
\end{axis}
\end{tikzpicture}
\label{fig:aleph600}
\end{minipage}
\caption{Predictive accuracies (\%) of \name{} versus \ale{} with 60s (left) and 600s (right) timeouts.}
\label{fig:aleph}
\end{figure}
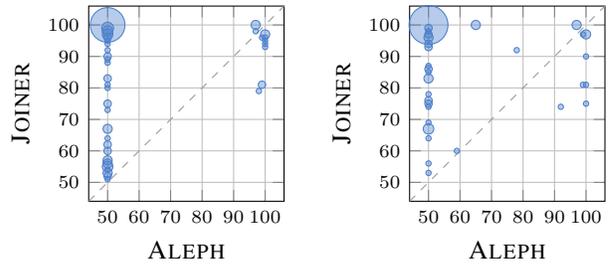

Figure \ref{fig:aleph} shows that \name{} outperforms \ale{}.
A McNeymar's test confirms ($p<0.01$) that \name{} outperforms \ale{} on 28/42 tasks with both 60s and 600s timeouts.
% a 60s timeout and on 28/42 tasks with a 600s timeout. 
\ale{} outperforms ($p < 0.01$) \name{} on 4/42 tasks with a 60s timeout and on 2/42 tasks with a 600s timeout. \name{} and \ale{} have similar performance on other tasks.

\ale{} struggles to learn recursive programs and therefore does not perform well on the \emph{string} tasks. 
\name{} also consistently surpasses \ale{} on tasks which do not require recursion. 
For instance, \name{} achieves 98\% average accuracy on the \emph{pharma} tasks while \ale{} has default accuracy (50\%). 
However, for \emph{zendo3}, \ale{} can achieve better accuracies (100\% vs 79\%) than \name{}. An optimal solution is not splittable and \name{} exceeds the timeout.
% and \ale{} achieves better accuracies (100\% vs 79\%).

Overall, the results suggest that the answer to \textbf{Q4} is that \name{} can outperform other approaches in terms of predictive accuracy.

\section{Conclusions and Limitations}
Learning programs with big rules is a major challenge.
% for current ILP systems. 
To address this challenge, we introduced an approach which learns big rules by joining small rules.
We implemented our approach in \name{}, which can learn optimal and recursive programs.
Our experiments on various domains show that \name{} can (i) learn splittable rules with more than 100 literals, and (ii) outperform existing approaches in terms of predictive accuracies.
% \name{} can learn large programs that other systems cannot.
% These results should help scale ILP to more complex applications.
%\paragraph{What worked and why?}

%\paragraph{Why should you care?}
%\ac{TODO}

%\paragraph{What did not work and why?}
%\ac{TODO}

%\paragraph{What can make this work better? Also think all off possible criticisms and mention them here.}

\subsection*{Limitations}

% \textbf{Scalability.}

\noindent
\textbf{Splittability.}
Our join stage builds splittable rules.
% , not non-splittable rules.
The body of a splittable rule is split into subsets which do not share body-only variables.
Future work should generalise our approach to split rules into subsets which may share body-only variables.
% Future work should allow joined rules to share body-only variables to learn big non-splittable rules. 

% \noindent
\textbf{Noise.} 
In the join stage, we search for conjunctions which entail some positive examples and no negative examples.
Our approach does not support noisy examples.
Hocquette et al. \shortcite{maxsynth} relax the LFF definition based on the minimal description length principle.
Future work should combine our approach with their approach to learn big rules from noisy data.

%% acks DC, FG, ML

\bibliographystyle{named}
\bibliography{ijcai24}

 \begin{appendices}
 
% This appendix contains supplementary material for the paper \emph{Learning Big Logical Rules by Joining Small Rules}.
% The outline is as follows.

% \begin{itemize}
% \item Section \ref{sec:bk} provides background on the terminology used in the paper.
% \item Section \ref{sec:encoding} describes our ASP encoding for the \emph{generate} stage of our algorithm.
% \item Section \ref{sec:proof} contains the proof of the
% correctness of our algorithm.
% \item Section \ref{sec:exp} describes our experiments, including sample solutions.
% \end{itemize}

\section{Terminology}
\label{sec:bk}
% \subsection{Logic Programming}
We assume familiarity with logic programming \cite{lloyd:book} but restate some key relevant notation. A \emph{variable} is a string of characters starting with an uppercase letter. A \emph{predicate} symbol is a string of characters starting with a lowercase letter. The \emph{arity} $n$ of a function or predicate symbol is the number of arguments it takes. An \emph{atom} is a tuple $p(t_1, ..., t_n)$, where $p$ is a predicate of arity $n$ and $t_1$, ..., $t_n$ are terms, either variables or constants. An atom is \emph{ground} if it contains no variables. A \emph{literal} is an atom or the negation of an atom. A \emph{clause} is a set of literals. A \emph{clausal theory} is a set of clauses. A \emph{constraint} is a clause without a positive literal. A \emph{definite} clause is a clause with exactly one positive literal. 
A \emph{hypothesis} is a set of definite clauses with the least Herbrand semantics.
% A \emph{program} is a set of definite clauses. 
We use the term \emph{program} interchangeably with hypothesis, i.e. a \emph{hypothesis} is a \emph{program}. 
A \emph{substitution} $\theta = \{v_1 / t_1, ..., v_n/t_n \}$ is the simultaneous replacement of each variable $v_i$ by its corresponding term $t_i$. 
A clause $c_1$ \emph{subsumes} a clause $c_2$ if and only if there exists a substitution $\theta$ such that $c_1 \theta \subseteq c_2$. 
A program $h_1$ subsumes a program $h_2$, denoted $h_1 \preceq h_2$, if and only if $\forall c_2 \in h_2, \exists c_1 \in h_1$ such that $c_1$ subsumes $c_2$. A program $h_1$ is a \emph{specialisation} of a program $h_2$ if and only if $h_2 \preceq h_1$. A program $h_1$ is a \emph{generalisation} of a program $h_2$ if and only if $h_1 \preceq h_2$. 

\section{Generate stage encoding}
\label{sec:encoding}
% \subsection{Generate} 
\label{generate}
Our encoding of the \emph{generate} stage is the same as \combo{} except we disallow splittable rules in hypotheses which are not recursive and do not contain predicate invention.
To do so, we add the following ASP program to the \emph{generate} encoding:
\begin{small}
\begin{verbatim}
:- not pi_or_rec, clause(C), splittable(C).

splittable(C) :-
     body_not_head(C,Var1),
     body_not_head(C,Var2),
     Var1 < Var2,
     not path(C,Var1,Var2).

path(C,Var1,Var2) :- path1(C,Var1,Var2).
path(C,Var1,Var2) :- 
    path1(C,Var1,Var3), 
    path(C,Var3,Var2).

path1(C,Var1,Var2) :-
     body_literal(C,_,_,Vars),
     var_member(Var1,Vars),
     var_member(Var2,Vars),
     Var1 < Var2,
     body_not_head(C,Var1),
     body_not_head(C,Var2).


%% we also disallow rules with multiple body 
literals when a body literal only has head variables
:- 
    not pi_or_rec, 
    body_literal(C,P1,_,Vars1), 
    body_literal(C,P2,_,Vars2), 
    (P1,Vars1) != (P2,Vars2), 
    head_only_vars(C,Vars1).

head_only_vars(C,Vars) :- 
    body_literal(C,_,_,Vars), 
    clause(C), 
    not has_body_var(C,Vars). 

has_body_var(C,Vars) :- 
    var_member(Var,Vars),
    body_not_head(C,Var).
    
body_not_head(C,V) :- 
    body_var(C,V), 
    not head_var(C,V).
\end{verbatim}
\end{small}

\section{\name{} Correctness}
\label{sec:proof}
We show the correctness of \name{}.
In the rest of this section, we consider a LFF input tuple $(E, B, \mathcal{H}, C)$ with $E=(E^+, E^-)$. We assume that a solution always exists:
\begin{assumption}[\textbf{Existence of a solution}] \label{existencesol}We assume there exists a solution $h \in \mathcal{H}_{C}$.
\end{assumption}
\noindent
For terseness, in the rest of this section we always assume Assumption \ref{existencesol} and, therefore, avoid repeatedly saying ``if a solution exists''.

We follow LFF \cite{popper} and assume the background knowledge does not depend on any hypothesis:
\begin{assumption}[\textbf{Background knowledge independence}] \label{independentbk}We assume that no predicate symbol in the body of a rule in the background knowledge appears in the head of a rule in a hypothesis. 
\end{assumption}
% In other words, we assume that the BK does not depend on any hypothesis.
\noindent
For instance, given the following background knowledge we disallow learning hypotheses for the target relations \emph{famous} or \emph{friend}:
\begin{center}
\begin{tabular}{l}
\emph{happy(A) $\leftarrow$ friend(A,B), famous(B)}\\
\end{tabular}
\end{center}
% \ch{do we need this?}

\noindent
As with \combo{} \cite{combo}, \name{} is correct when the hypothesis space only contains decidable programs (such as Datalog programs), i.e. when every program is guaranteed to terminate:
\begin{assumption}[\textbf{Decidable programs}]\label{decidable}
We assume the hypothesis space only contains decidable programs.
\end{assumption}
% which also holds for \popper{} \cite{popper}, 
\noindent
When the hypothesis space contains arbitrary definite programs, the results do not hold because, due to their Turing completeness, checking entailment of an arbitrary definite program is semi-decidable \cite{tarnlund:hornclause}, i.e. testing a program might never terminate.
% \ch{perhaps number the assumptions to refer back to them}

\name{} follows a generate, test, join, combine, and constrain loop. We show the correctness of each of these steps in turn.
% We follow the proof of the correctness of \combo{} \cite{combo} and we prove the correctness of \name{} in three steps. 
% We first assume that there is no constrain stage, i.e. \name{} does not apply constraints.
% With this assumption, we show that the generate, test, join and combine stages return an optimal solution. 
% We then show that the constrain stage is optimally sound in that it never prunes optimal solutions from the hypothesis space.
We finally use these results to prove the correctness of \name{} i.e. that \name{} returns an optimal solution.

\subsection{Preliminaries}
We introduce definitions used throughout this section.
We define a splittable rule:
\begin{definition}[\textbf{Splittable rule}]
A rule is splittable if its body literals can be partitioned into two non-empty sets such that the body-only variables (a variable in the body of a rule but not the head) in the literals of one set are disjoint from the body-only variables in the literals of the other set.
A rule is non-splittable if it is not splittable.
\label{def:splittablerule_appendix}
\end{definition}

\noindent
We define a splittable program:
\begin{definition}
[\textbf{Splittable program}]
A program is \emph{splittable} if and only if it has exactly one rule and this rule is splittable.
A program is non-splittable when it is not splittable.
\label{def:splittable_appendix}
\end{definition}

\noindent
The least Herbrand model $M(P)$ of the logic program $P$ is the set of all ground atomic logical consequences of $P$,
The least Herbrand model $M_{B}(P) = M(P \cup B)$ of $P$ given the logic program $B$ is the the set of all ground atomic logical consequences of $P \cup B$. 
In the following, we assume a program $B$ denoting background knowledge and concisely note $M_{B}(P)$ as $M(P)$.

% \ac{@CH, I would move these definitions just before they are used as I would skip this section when reading a paper.}

We define a conjunction:
\begin{definition}[\textbf{Conjunction}]
A conjunction is a set of programs with the same head literal. 
The logical consequences of a conjunction $c$ is the intersection of the logical consequences of the programs in it: $M(c) = \cap_{p \in c} M(p)$. 
The cost of a conjunction $c$ is the sum of the cost of the programs in it: $cost(c) = \sum_{p\in c} cost(p)$.
\label{conjunction}
\end{definition}

\noindent
Conjunctions only preserve the semantics with respect to their head predicate symbols.
We, therefore, reason about the least Herbrand model restricted to a predicate symbol:
\begin{definition}[\textbf{Restricted least Herbrand model}]
Let $P$ be a logic program and $f$ be a predicate symbol. 
Then the least Herbrand model of $P$ restricted to $f$ is $M(P,f) = \{ a \in M(P) \,|\, \text{the predicate symbol of $a$ is } f\}$.
\end{definition}

\noindent
We define an operator $\sigma: 2^{\mathcal{H}} \mapsto \mathcal{H}$ which maps a conjunction to a program such that (i) for any conjunction $c$ with head predicate symbol $f$, $M(c, f)=M(\sigma(c), f)$,
% \ac{h is overused as it often means hypothesis, I suggest using a different symbol}
% \ac{@CH this statement does not hold as we need to rename the predicate symbols in a conjunction. We do not need to say this statement.}.
and (ii) for any conjunctions $c_1$ and $c_2$ such that $size(c_1) < size(c_2)$, $size(\sigma(c_1)) < size(\sigma(c_2))$.
In the following, when we say that a conjunction  \emph{$c$ is in a program $h$}, or $c \subseteq h$, we mean \emph{$\sigma(c) \subseteq h$}.

For instance, let $\sigma$ be the operator which, given a conjunction $c$ with head predicate symbol $f$ of arity $a$, returns the program obtained by (i) for each program $p \in c$, replacing each occurrence of $f$ in $p$ by a new predicate symbol $f_i$, and (ii) adding the rule $f(x_1, \dots, x_a) \leftarrow f_1(x_1, \dots, x_a), \dots, f_n(x_1, \dots, x_a)$ where $x_i, \dots, x_a$ are variables.
\begin{example}
Consider the following conjunction:
\[\begin{array}{l}
c_1 = \left\{
    \begin{array}{l}
     \left\{
    \begin{array}{l}
\emph{zendo(S) $\leftarrow$ piece(S,B), blue(B)}\\
    \end{array}
    \right\}\\
    \left\{
    \begin{array}{l}
\emph{zendo(S) $\leftarrow$ piece(S,R), red(R)}\\
    \end{array}
    \right\}
    \end{array}
    \right\}
    \end{array}
\]
Then $\sigma(c_1)$ is:
\[
\begin{array}{l}
\left\{
    \begin{array}{l}
\emph{zendo(S) $\leftarrow$ zendo$_1$(S),zendo$_2$(S)}\\
\emph{zendo$_1$(S) $\leftarrow$ piece(S,B), blue(B)}\\
\emph{zendo$_2$(S) $\leftarrow$ piece(S,R), red(R)}\\
\end{array}
\right\}
\end{array}
\]

\noindent
Consider the following conjunction:
\[\begin{array}{l}
c_2 = \left\{
    \begin{array}{l}
     \left\{
    \begin{array}{l}
\emph{f(S) $\leftarrow$ head(S,1)}\\
\emph{f(S) $\leftarrow$ tail(S,T),f(T)}\\
    \end{array}
    \right\}\\
    \left\{
    \begin{array}{l}
\emph{f(S) $\leftarrow$ head(S,2)}\\
\emph{f(S) $\leftarrow$ tail(S,T),f(T)}\\
    \end{array}
    \right\}
    \end{array}
    \right\}
    \end{array}
\]
Then $\sigma(c_2)$ is:
\[
\begin{array}{l}
\left\{
    \begin{array}{l}
\emph{f(S) $\leftarrow$ f$_1$(S),f$_2$(S)}\\
\emph{f$_1$(S) $\leftarrow$ head(S,1)}\\
\emph{f$_1$(S) $\leftarrow$ tail(S,T),f$_1$(T)}\\
\emph{f$_2$(S) $\leftarrow$ head(S,2)}\\
\emph{f$_2$(S) $\leftarrow$ tail(S,T),f$_2$(T)}\\
\end{array}
\right\}
\end{array}
\]
% \ac{I would use the zendo example from section 1 of the paper}
\end{example}
% \ac{|@CH, add a recursive example}

\noindent
The generate stage of \combo{} generates non-separable programs.
We recall the definition of a separable program:
\begin{definition}
[\textbf{Separable program}]
A program $h$ is \emph{separable} when (i) it has at least two rules, and (ii) no predicate symbol in the head of a rule in $h$ also appears in the body of a rule in $h$. A program is \emph{non-separable} when it is not separable.
\end{definition}

\subsection{Generate and Test}
\noindent
We show that \name{} can generate and test every non-splittable non-separable program:
\begin{proposition}[\textbf{Correctness of the generate and test stages}]
\label{lem_generate}
\name{} can generate and test every non-splittable non-separable program.
\end{proposition}
\begin{proof}
Cropper and Hocquette \shortcite{combo} show (Lemma 1) that \combo{} can generate and test every non-separable program under Assumption \ref{decidable}.
% the generate encoding of \combo{} has a model for every program in the hypothesis space.
The constraint we add to the generate encoding (Section \ref{generate}) prunes splittable programs.
% a subset of all non-separable programs.
Therefore \name{} can generate and test every non-splittable non-separable program.
\end{proof}

\subsection{Join}
Proposition \ref{lem_generate} shows that \name{} can generate and test every non-splittable non-separable program. 
However, the generate stage cannot generate splittable non-separable programs.
To learn a splittable non-separable program, \name{} uses the join stage to join non-splittable programs into splittable programs.
To show the correctness of this join stage, we first show that the logical consequences of a rule which body is the conjunction of the body of two rules $r_1$ and $r_2$ is equal to the intersection of the logical consequences of $r_1$ and $r_2$: 

\begin{lemma}
% [\textbf{Conjunction of non-splittable rules}]
\label{eq}
Let 
$r_1 = (g \leftarrow L_1)$, 
$r_2 = (g \leftarrow L_2)$, 
and $r = (g \leftarrow L_1, L_2)$ be three rules, such that the body-only variables of $r_1$ and $r_2$ are distinct.
% and $c_1, c_2,$ and $c$ be three conjunctions such that 
% $c = c_1 \cup c_2$,
% $M(r_1) = M(c_1)$, 
% $M(r_2) = M(c_2)$,
Then $M(r) = M(r_1) \cap M(r_2)$.
\end{lemma}
% \begin{proof}
% \begin{align*}
% M(c) &= \cap_{p \in c} M(p)\\
% &= (\cap_{p \in c_1} M(p)) \cap (\cap_{p \in c_2} M(p))\\
% &= M(c_1) \cap M(c_2)\\
% &= M(r_1) \cap M(r_2)\\
% \end{align*}
% Then $\sigma(c)$ is:
% \[
% \begin{array}{l}
% \left\{
%     \begin{array}{l}
% \emph{$g \leftarrow g_{1,1},\dots, g_{1,l}, g_{2,1},\dots, g_{2,m}$}\\
% \end{array}
% \right\}
% \end{array}
% \]
% where $g_{1,i} \leftarrow M_i$ with $M_i \subseteq L_1$ and $\cup_{i} M_i = L_1$\\
% and $g_{2,i} \leftarrow M_j$ with $M_j \subseteq L_2$ and $\cup_{j} M_2= L_2$\\
% From the correctness of the unfold operator, we have:
% Then $\sigma(c)$ is equivalent to:
% \[
% \begin{array}{l}
% \left\{
%     \begin{array}{l}
% \emph{$g \leftarrow L_1, L_2$}\\
% \end{array}
% \right\}
% \end{array}
% \]
% \end{proof}

\begin{proof}
% a in mc implies a in mr
We first show that if $a\in~M(r)$ then $a~\in M(r_1)~\cap~M(r_2)$.
The rule $r$ specialises $r_1$ which implies that $M(r)\subseteq~M(r_1)$ and therefore $a \in M(r_1)$.
Likewise, $r$ specialises $r_2$ which implies that $M(r) \subseteq M(r_2)$ and therefore $a \in M(r_2)$. 
Therefore $a \in M(r_1) \cap M(r_2)$.
% Since the body-only variables of $r_1$ and $r_2$ are distinct the logical consequences are independent.
% \ac{@AC, confused myself}

We now show that if $a \in M(r_1) \cap M(r_2)$ then $a \in M(r)$. 
Since $a \in M(r_1)$ then $L_1 \lor \neg g \models a$. 
Similarly, since $a \in M(r_2)$ then $L_2 \lor \neg g \models a$. 
Then $(L_1 \lor \neg g) \land (L_2 \lor \neg g) \models a$ by monotonicity, and $a \in M(r)$. 
% \ch{still very unsure about how to prove that}
% \ac{it also does not seem correct as a is defined with the predicate symbol of g}
\end{proof}

\noindent
With this result, we show that any rule is equivalent to a conjunction of non-splittable programs:
\begin{lemma}[\textbf{Conjunction of non-splittable programs}]
\label{lem_splittable}
% Let $r=(h,b)$ be a rule with head $h = f(A,B,...)$ of arity $a$ and with body $b$. Then $r$ can be decomposed as $h\leftarrow inv_1(A,B,...), inv_2(A,B,...), ...$ where each $inv_i$ is a new predicate symbol not appearing in $r$ and with arity $a$ and each $inv_i(A,B,...)$ is defined with a non-splittable rule with literals from $b$ in its body.
% Then $b = \bigcup_{i=1}^n b_{i}$ where each rule with head $h$ and body $b_{i}$ is a non-splittable rule.
Let $r$ be a rule. Then there exists a conjunction $c$ of non-splittable programs such that $M(r) = M(c)$.
\label{rules}
\end{lemma}

% \begin{proof}
% \ch{NEEEW}
% We prove the result by induction on the number of body literals $m$ in the body of $r$. 

% For the base case, if $m=1$ then $r$ is non-splittable. 
% Let $c$ be the conjunction $\{\{r\}\}$. 
% Then $M(r) = M(c)$.
% % \ac{M(c) is really confusing as c is a set of sets. I know we redefine M(c) for a conjunction but it is confusing. I am unsure what to do.}

% For the inductive case, assume the claim holds for rules with $m$ body literals.
% % We show the claim holds for rules with $m+1$ body literals.
% Let $r$ be a rule with $m+1$ body literals.
% Either $r$ is (i) non-splittable, or (ii) splittable.
% For case (i), assume $r$ is non-splittable. Let $c$ be the conjunction $\{\{r\}\}$. Then $M(r) = M(c)$.
% For case (ii), assume $r$ is splittable. 
% Then, by Definition \ref{def:splittablerule}, its body literals can be partitioned into two non-empty sets $L_1$ and $L_2$ such that the body-only variables in $L_1$ are disjoint from the body-only variables in $L_2$.
% Both $L_1$ and $L_2$ have fewer than $m$ literals.
% Let $a$ be the head literal of $r$.
% Let $r_1 = (a_1 \leftarrow L_1)$, $r_2 = (a_2 \leftarrow L_2)$, $r_3 = (a \leftarrow a_1, a_2)$. 
% By the induction hypothesis, there exists a conjunction $c_1$ of non-splittable programs such that $M(r_1)=M(c_1)$ and a conjunction $c_2$ of non-splittable programs such that $M(r_2)=M(c_2)$.
% Let $c = r_1 \cup r_2 \cup r_3$.
% $M(r)=M_{\{a\}}(c)$ from the correctness of the unfold operator.
% but these rules are not splittable and c is not a conjunction
% \end{proof}

\begin{proof}
We prove the result by induction on the number of body literals $m$ in the body of $r$. 

For the base case, if $m=1$ then $r$ is non-splittable. 
Let $c$ be the conjunction $\{\{r\}\}$. 
Then $M(r) = M(c)$.
% \ac{M(c) is really confusing as c is a set of sets. I know we redefine M(c) for a conjunction but it is confusing. I am unsure what to do.}

For the inductive case, assume the claim holds for rules with $m$ body literals.
% We show the claim holds for rules with $m+1$ body literals.
Let $r$ be a rule with $m+1$ body literals.
Either $r$ is (i) non-splittable, or (ii) splittable.
For case (i), assume $r$ is non-splittable. Let $c$ be the conjunction $\{\{r\}\}$. Then $M(r) = M(c)$.
For case (ii), assume $r$ is splittable. 
Then, by Definition \ref{def:splittablerule_appendix}, its body literals can be partitioned into two non-empty sets $L_1$ and $L_2$ such that the body-only variables in $L_1$ are disjoint from the body-only variables in $L_2$.
Both $L_1$ and $L_2$ have fewer than $m$ literals.
Let $a$ be the head literal of $r$.
Let $r_1 = (a \leftarrow L_1)$ and $r_2 = (a \leftarrow L_2)$. 
By the induction hypothesis, there exists a conjunction $c_1$ of non-splittable programs such that $M(r_1)=M(c_1)$ and a conjunction $c_2$ of non-splittable programs such that $M(r_2)=M(c_2)$.
Let $c = c_1 \cup c_2$. 
Then $c$ is a conjunction because the programs in $c_1$ and $c_2$ have the same head predicate symbol.
From Lemma \ref{eq}, $M(r)=M(r_1) \cap M(r_2)$ and therefore $M(r)=M(c_1) \cap M(c_2)=M(c)$ which completes the proof.
\end{proof}

\begin{myexample}[\textbf{Conjunction of non-splittable programs}] \label{ex1}
Consider the rule $r$:
\[
    \begin{array}{l}
    r = \left\{
    \begin{array}{l}
\emph{f(A) $\leftarrow$ head(A,1),tail(A,B),head(B,3)}\\
    \end{array}
    \right\}\\
    \end{array}
\]
Consider the non-splittable programs:
\[
    \begin{array}{l}
    p_1 = \left\{
    \begin{array}{l}
\emph{f(A) $\leftarrow$ head(A,1)}\\
    \end{array}
    \right\}\\
    p_2 = \left\{
    \begin{array}{l}
\emph{f(A) $\leftarrow$ tail(A,B),head(B,3)}\\
    \end{array}
    \right\}
    \end{array}
\]
Then $M(r) = M(c)$ where $c$ is the conjunction $c=p_1 \cup p_2$.
\end{myexample}

\noindent
We show that any program is equivalent to a conjunction of non-splittable programs:
\begin{lemma}[\textbf{Conjunction of non-splittable programs}]
\label{lem_splittable_program}
Let $h$ be a program. Then there exists a conjunction $c$ of non-splittable programs such that $M(h) = M(c)$.
\end{lemma}
\begin{proof}
$h$ has either (i) at least two rules, or (ii) one rule.
For case (i), if $h$ has at least two rules, then $h$ is non-splittable. Let $c$ be the conjunction $\{\{h\}\}$. Then $M(h) = M(c)$.
For case (ii), Lemma \ref{rules} shows there exists a conjunction $c$ of non-splittable programs such that $M(h) = M(c)$.
\end{proof}
% \ch{do we also need something about the size of the conjunction?}

\noindent
The join stage takes as input \emph{joinable} programs. We define a \emph{joinable} program:
\begin{definition}[\textbf{Joinable program}]
 A program is \emph{joinable} when it (i) is non-splittable, (ii) is non-separable, (iii) entails at least one positive example, and (iv) entails at least one negative example.
 % ($tp > 0$).
\end{definition}

\noindent
The combine stage takes as input \emph{combinable} programs. We define a \emph{combinable} program:
\begin{definition}[\textbf{Combinable program}]
A program is \emph{combinable} when it (i) is non-separable, (ii) entails at least one positive example, and (iii) entails no negative example.
\end{definition}

\noindent
We show that any combinable program is equivalent to a conjunction of non-splittable non-separable programs which each entail at least one positive example:
\begin{lemma}[\textbf{Conjunction of programs}]\label{lem_opti_separable}
Let $h$ be combinable program. Then there exists a conjunction $c$ of programs such that $M(c)=M(h)$ and each program in $c$ (i) is non-splittable, (ii) is non-separable, and (iii) entails at least one positive example.
\end{lemma}

\begin{proof}
From Lemma \ref{lem_splittable_program}, there exists a conjunction $c$ of non-splittable programs such that $M(c)=M(h)$.
For contradiction, assume some program $h_i \in c$ is either (i) separable, or (ii) does not entail any positive example. 
For case (i), assume that $h_i$ is separable. 
Then $h_i$ has at least two rules and therefore cannot be non-splittable, which contradicts our assumption.
For case (ii), assume $h_i$ does not entail any positive example.
Then $c$ does not entail any positive example and $h$ does not entail any positive example. 
Then $h$ is not a combinable program, which contradicts our assumption.
Therefore, each $h_i \in c$ (i) is non-splittable, (ii) is non-separable, and (iii) entails at least one positive example.
\end{proof}

\noindent
The join stage returns non-subsumed conjunctions. A conjunction is subsumed if it entails a subset of the positive examples entailed by a strictly smaller conjunction:
% entails a subset of the positive examples entailed by a strictly smaller conjunction 
\begin{definition}[\textbf{Subsumed conjunction}] Let $c_1$ and $c_2$ be two conjunctions which do not entail any negative examples,
$c_1 \models E^{+}_1$,
$c_2 \models E^{+}_2$,
$E^{+}_2 \subseteq E^{+}_1$,
$size(c_1) < size(c_2)$. Then $c_2$ is subsumed by $c_1$.
\end{definition}

% \ch{does the following result help to prove the next proposition (correctness of the join stage)? the join stage only takes as input programs which cover at least one neg ex and returns conjunctions of at least two programs}
% \ch{should 'joinable' (def 6) also cover some negative examples?}
\begin{lemma}[\textbf{Conjunction of joinable programs}]\label{lem_splittable_separable}
Let $h$ be a non-subsumed splittable combinable program. Then there exists a conjunction $c$ of joinable programs such that $M(c)=M(h)$.
\end{lemma}
\begin{proof}
Since $h$ is splittable, $h$ has a single rule which body literals can be partitioned into two non-empty sets $L_1$ and $L_2$ such that the body-only variables in $L_1$ are disjoint from the body-only variables in $L_2$. Let $g$ be the head literal of the rule in $h$. Let $r_1 = (g \leftarrow L_1)$ and $r_2 = (g \leftarrow L_2)$. 
From Lemma \ref{lem_opti_separable}, there exists two conjunctions $c_1$ and $c_2$ such that $M(c_1)=M(r_1)$ and $M(c_2)=M(r_2)$ and each program in $c_1 \cup c_2$ (i) is non-splittable, (ii) is non-separable, and (iii) entails at least one positive example.
Let $c = c_1 \cup c_2$. Then $c$ is a conjunction because the programs in $c_1$ and $c_2$ have the same head predicate symbol. 
From Lemma \ref{eq}, $M(r)=M(r_1) \cap M(r_2)$ and therefore $M(r) = M(c_1) \cap M(c_2) = M(c)$.
Therefore there exists a conjunction $c$ of at least two programs such that $M(c)=M(h)$ and each program in $c$ (i) is non-splittable, (ii) is non-separable, and (iii) entails at least one positive example.
Assume some $h_i \in c$ does not cover any negative example.
Then the conjunction $c' = \{\{h_i\}\}$ does not entail any negative example. Since $c$ has at least two elements, then $c' \subset c$, and $c$ is subsumed by $c'$ which contradicts our assumption. Therefore, each $h_i \in c$ entails some negative examples and is joinable.
\end{proof}

\noindent
We show that the join stage returns all non-subsumed splittable combinable programs:
\begin{proposition}[\textbf{Correctness of the join stage}]
\label{lem_join}
% Let $h$ be a splittable combinable program. Given the set of all joinable programs, the join stage returns a conjunction $c$ such that $M(c) = M(h)$.\ac{OLD}
Given all joinable programs, the join stage returns all non-subsumed splittable combinable programs.
% \ac{NEW}
% s all splittable combinable programs.
\end{proposition}
\begin{proof}
Assume the opposite, i.e. that there is a non-subsumed splittable combinable program $h$ not returned by the join stage.
As $h$ is non-subsumed splittable combinable, by Lemma \ref{lem_splittable_separable}, there exists a conjunction $c$ of joinable programs such that $M(c)=M(h)$. 
As $h$ is non-subsumed, $c$ is not subsumed by another conjunction.
The join stage uses a SAT solver to enumerate maximal satisfiable subsets \cite{DBLP:journals/jar/LiffitonS08} corresponding to the subset-maximal coverage of conjunctions.
As we have all joinable programs, the solver will find every conjunction of joinable programs not subsumed by another conjunction.
Therefore, the solver finds $c$ and the join stage returns $h$, which contradicts our assumption.

% which implies the join stage does not return a conjunction $c$ such that $M(c) = M(h)$.
% For contradiction, assume the opposite, which implies the join stage does not return a conjunction $c$ such that $M(c) = M(h)$.
% there exists a splittable combinable program $h$ such that the join stage does not return a program equivalent to $h$.

% For all $h_i \in c$, then $h_i$.
% Since $h$ is combinable, $h$ entails at least one positive example. 
% There are finitely many joinable programs. 
% Moreover, since $h$ is combinable, $c$ is a model for the SAT encoding.
% The SAT solver finds all models that entails at least one positive example and no negative examples.
% Therefore the SAT solver finds $c$.
% The assumption cannot hold and the join stage returns a conjunction $c$ such that $M(c) = M(h)$.
% \ch{slightly unclear, needs rephrasing, as we find c not h}
% \ac{ASP to SAT?}
\end{proof}

\subsection{Combine}

% Let $\sigma$ be an operator which maps subsets of $\mathcal{H}$ to programs, i.e. sets of programs to programs such that $M(P) = M(\sigma(P))$.

% We define an operator $\sigma: 2^{\mathcal{H}} \mapsto \mathcal{H}$ which maps subsets of $\mathcal{H}$ to programs \ac{NEW???}

% In the following, we say \emph{c is in a program} to mean \emph{$\sigma(c) \in P$}.

% \ch{need to show such an operator exists}
% \ac{we need to define it}
% \ch{say something about the size of conjunctions?}
% \ac{operator should probably go in with the definition of a conjunction}

\noindent
We first show that a subsumed conjunction cannot be in an optimal solution:
\begin{proposition}[\textbf{Subsumed conjunction}]
\label{prop_subsumed}
Let $c_1$ and $c_2$ be two conjunctions such that $c_2$ is subsumed by $c_1$,
% which do not entail any negative examples,
% $c_1 \models E^{+}_1$,
% $c_2 \models E^{+}_2$,
% $E^{+}_2 \subseteq E^{+}_1$,
% $size(c_1) < size(c_2)$,
$h$ be a solution, 
and $\sigma(c_2) \subseteq h$.
Then $h$ cannot be optimal.
\end{proposition}
\begin{proof}
% Assume there exists an optimal solution $h_o$ and a hypothesis $h$ such that $h_o = c_2 \cup h$.
Assume the opposite, i.e. $h$ is an optimal solution.
% Assume there exists an optimal solution $h_o$ and a hypothesis $h$ such that $h_o = h \cup \sigma{}(c_2)$.
% We show that $h'_o = c_1 \cup h$ is a solution.
We show that $h' = (h \setminus \sigma(c_2)) \cup \sigma(c_1)$ is a solution.
Since $h$ is a solution, it does not entail any negative examples.
Therefore, $h \setminus \sigma(c_2)$ does not entail any negative examples. 
Since $c_1$ does not entail any negative examples, then $h'$ does not entail any negative examples.
Since $h$ is a solution, it entails all positive examples. 
Since $E^{+}_2 \subseteq E^{+}_1$, $h'$ also entails all positive examples.
Therefore $h'$ is a solution.
Since $size(c_1) < size(c_2)$ then $size(h') < size(h)$. 
Then $h$ cannot be optimal, which contradicts our assumption.
\end{proof}

\begin{corollary}
Let $c_1$ and $c_2$ be two conjunctions which do not entail any negative examples,
% and entail the positive examples $E^{+}_1$ and $E^{+}_2$ respectively, and
% which entail the positive examples 
$c_1 \models E^{+}_1$,
$c_2 \models E^{+}_2$,
% and $E^{+}_2$ respectively and no negative example, 
% such that
$E^{+}_2 \subseteq E^{+}_1$, 
and $size(c_1) < size(c_2)$. 
Then $c_2$ cannot be in an optimal solution.
\label{coro:subsumed}
\end{corollary}

\begin{proof}
Follows from Proposition \ref{prop_subsumed}.
\end{proof}

% \noindent
% We show the correctness of the combine stage. 
\noindent
With this result, we show the correctness of the combine stage:
\begin{proposition}[\textbf{Correctness of the combine stage}] \label{prop_combine}
Given all non-subsumed combinable programs, the combine stage returns an optimal solution.
\end{proposition}
\begin{proof}
Cropper and Hocquette [\citeyear{combo}] show that given all combinable programs, the combine stage returns an optimal solution under Assumption \ref{independentbk}. Corollary \ref{coro:subsumed} shows that subsumed conjunctions cannot be in an optimal solution. 
Therefore, the combine stage of \name{} returns an optimal solution given all non-subsumed combinable programs.
\end{proof}

% \noindent
% We show that \name{} (with only the generate, test, join and combine stages) returns a solution:
% \begin{lemma}[\textbf{Correctness of \name{} without constraints}]
% \label{lem_gen_one_opt}
% \name{} returns a solution.
% \end{lemma}
% \begin{proof}
% For contradiction, assume the opposite, i.e. \name{} does not return a solution.
% This assumption implies that \name{} (i) returns a hypothesis that is not a solution, or (ii) does not return a hypothesis.
% Case (i) cannot hold because as an output of the combine stage $h$ is complete and consistent so is a solution. 
% Case (ii) cannot hold because 
% Lemma \ref{lem_generate} shows that \name{} can generate and test every non-splittable non-separable program, of which there are finitely many.
% Lemma \ref{lem_join} shows that the join stage can build every splittable combinable program, of which there are finitely many.
% Therefore, \name{} can generate every combinable program.
% Given the set of all combinable programs, the combine stage returns a solution \cite{combo}.
% By Assumption \ref{existencesol}, a solution exists.
% Therefore \name{} returns a solution.
% \end{proof}

% \noindent
% We show that \name{} without constraints returns an optimal solution.
% \begin{proposition}[\textbf{Correctness of \name{} without constraints}]
% \label{prop1}
% \name{} without constraints returns an optimal solution.
% \end{proposition}
% \begin{proof}
% Follows from Lemma \ref{lem_gen_one_opt} and that \name{} uses iterative deepening over the program size to guarantee optimality.
% \end{proof}

\subsection{Constrain}

% In the previous section, we showed that \name{} finds an optimal solution when it does not apply any constraints.
In this section, we show that the constrain stage never prunes optimal solutions from the hypothesis space.
\name{} builds specialisations constraints from non-splittable non-separable programs that do not entail (a) any positive example, or (b) any negative example. 
% Moreover, \name{} does not build subsumed conjunctions (constraint (c)).

% \noindent
We show a result for case (a). If a hypothesis does not entail any positive example, \name{} prunes all its specialisations, as they cannot be in a conjunction in an optimal solution:
\begin{proposition}[\textbf{Specialisation constraint when tp=0}]
\label{prop:totincomplete_appendix}
% Let $h_1$ be a hypothesis that does not entail any positive example 
% and $h_2$ be a specialisation of $h_1$.
% Then $h_2$ cannot be in a conjunction $c$ such that $\sigma(c)$ is in an optimal solution.
% \ac{NEW}
% Let $h_1$ be a hypothesis that does not entail any positive example,
% $h_2$ be a specialisation of $h_1$, 
% $h_o$ be an optimal solution, 
% $c$ be a conjunction,
% and $\sigma(c) \subseteq{} h_o$.
% Then $h_2 \not\in c$.
% NEW2
% Let $h_1$ be a hypothesis that does not entail any positive example,
% $h_2$ be a specialisation of $h_1$, 
% $h$ be a hypothesis,
% $c$ be a conjunction,
% and $\sigma(c) \subseteq{} h_o$.
% Then $h_2 \not\in c$.
% NEW3 
Let $h_1$ be a hypothesis that does not entail any positive example,
$h_2$ be a specialisation of $h_1$, 
$c$ be a conjunction,
$h_2 \in c$,
$h$ be a solution,
and $\sigma{}(c) \subseteq{} h$.
Then $h$ is not an optimal solution.
\end{proposition}
\begin{proof}
% OLD
% Assume the opposite, i.e. there exists an optimal solution $h_o$ and a conjunction $c$ such that $h_2 \in c$ and $\sigma{}(c) \subseteq h_o$. 
% NEW
% Assume the opposite, i.e. $h_2 \in c$.
Assume the opposite, i.e. $h$ is an optimal solution.
We show that $h' = h \setminus \sigma{}(c)$ is a solution.
% \ac{<--- @CH, when the assumptions are in the proposition, this sentence does not make sense}
Since $h$ is a solution, it does not entail any negative example. Then $h'$ does not entail any negative example.
Since $h_1$ does not entail any positive example and $h_2$ specialises $h_1$ then $h_2$ does not entail any positive example. Then $c$ does not entail any positive example. 
Since $h$ entails all positive examples and $c$ does not entail any positive example, then $h'$ entails all positive examples.
Since $h'$ entails all positive examples and no negative examples, it is a solution.
Moreover, $size(h') < size(h)$.
Then $h$ cannot be optimal, which contradicts our assumption.
\end{proof}

\begin{corollary}
% [\textbf{Specialisation constraint when tp=0}]
Let $h_1$ be a hypothesis that does not entail any positive example
and $h_2$ be a specialisation of $h_1$.
Then $h_2$ cannot be in a conjunction in an optimal solution.
\label{cor_totincomplete}
\end{corollary}
\begin{proof}
Follows from Proposition \ref{prop:totincomplete_appendix}.
\end{proof}

% \begin{myexample}[\textbf{Specialisation constraint when tp=0}]
% % [
% % \textbf{Specialisations of totally incomplete programs}]
%     Consider the hypothesis:
% \[
%     \begin{array}{l}
%     h_1 = \left\{
%     \begin{array}{l}
%     \emph{f(A,B) $\leftarrow$ head(A,B)}
%     \end{array}
%     \right\}
%     \end{array}
% \]
% If $h_1$ does not entail any positive example, we can prune its specialisations, such as the $h_2$, as they cannot be in a conjunction of an optimal solution.
% \[
%     \begin{array}{l}
%     h_2 = \left\{
%     \begin{array}{l}
%     \emph{f(A,B) $\leftarrow$ head(A,B), even(B)}
%     \end{array}
%     \right\}
%     \end{array}
% \]
% For instance, $h_3$ cannot be an optimal solution:
% \[
%     \begin{array}{l}
%     h_3 = \left\{
%     \begin{array}{l}
%     \emph{f(A,B) $\leftarrow$ f1(A,B),f2(A,B)}\\
%     \emph{f1(A,B) $\leftarrow$ head(A,B), even(B)}\\
%     \emph{f2(A,B) $\leftarrow$ last(A,C), odd(C)}\\
%     \emph{f(A,B) $\leftarrow$ tail(A,C), head(C,B)}
%     \end{array}
%     \right\}
%     \end{array}
% \]
% \noindent
% It is important to note that case (a) does not prohibit specialisations of $h_1$ from being in a non-splittable rule of an optimal solution, such as:
% \[
%     \begin{array}{l}
%     h_3 = \left\{
%     \begin{array}{l}
%     \emph{f(A,B) $\leftarrow$ head(A,B), odd(B)}\\
%     \emph{f(A,B) $\leftarrow$ tail(A,C), f(C,B)}\\
%     \end{array}
%     \right\}
%     \end{array}
% \]
% \end{myexample}

\noindent
We show a result for case (b). If a hypothesis does not entail any negative example, \name{} prunes all its specialisations, as they cannot be in a conjunction in an optimal solution:
\begin{proposition}[\textbf{Specialisation constraint when fp=0}]
\label{prop:consistent_appendix}
% OLD
% Let $h_1$ be a hypothesis that does not entail any negative example and $h_2$ be a specialisation of $h_1$.
% Then $h_2$ cannot be in a conjunction $c$ such that $\sigma(c)$ is in an optimal solution.
% Let $h_1$ be a hypothesis that does not entail any negative example, 
% and $h_2$ be a specialisation of $h_1$,
% Then $h_2$ cannot be in a conjunction $c$ such that $\sigma(c)$ is in an optimal solution.
Let $h_1$ be a hypothesis that does not entail any negative example,
$h_2$ be a specialisation of $h_1$, 
$c$ be a conjunction,
$h_2 \in c$,
$h$ be a solution,
and $\sigma(c) \subseteq{} h$.
Then $h$ is not an optimal solution.
\end{proposition}
\begin{proof}
% \ac{OLD Assume the opposite, i.e. there exists an optimal solution $h_o$, a conjunction $c$, and a hypothesis $h_3$ such that $h_2 \in c$ and $h_o = h_3 \cup \sigma{}(c)$.}
% \ac{NEW: Assume the opposite, i.e. $h$ is an optimal solution.}
Assume the opposite, i.e. $h$ is an optimal solution.
Let $c'=(c \setminus \{h_2\}) \cup \{h_1\}$ and $h'=(h \setminus \sigma{}(c)) \cup \sigma{}(c')$.
We show that $h'$ is a solution.
Since $h_1$ does not entail any negative example, then $c'$ does not entail any negative example.
Since $h$ is a solution, it does not entail any negative examples.
Then $h'$ do not entail any negative example.
Since $h_2$ specialises $h_1$, then $h_1$ entails more positive examples than $h_2$. Then $c'$ entails more positive examples than $c$, and $h'$ entails more positive examples than $h$.
Since $h$ entails all positive examples then $h'$ entails all positive examples.
Since $h'$ entails all positive examples and no negative examples, it is a solution.
Since $h_2$ specialises $h_1$ then $size(h_1) < size(h_2)$ and $size(c') < size(c)$.
Therefore, $size(h') < size(h)$.
Then $h$ cannot be optimal, which contradicts our assumption.
% (i) an optimal solution or (ii) in an optimal splittable solution. For case (i), assume $h_2$ is an optimal solution. Since $h_2$ is complete and $h_2$ specialises $h_1$ then $h_1$ is complete. Since $h_1$ is complete and consistent then it is a solution. Since $h_2$ specialises $h_1$ then $size(h_1) < size(h_2)$.
% Therefore, $size(h_1) < size(h_2)$ so $h_2$ is not an optimal solution and case (i) cannot hold.
% For case (ii), assume $h_2$ is in an optimal splittable solution $h_o$.
% Then $h_o = h_2 \cup h_3$ where $h_3 \neq \emptyset$.
% Let $h' = h_1 \cup h_3$. 
% Since $h_o$ is complete and $h_2$ specialises $h_1$ then $h'$ is complete.
% Since $h_o$ is consistent then $h_3$ is consistent. Since $h_o$ is splittable and $h_2$ specialises $h_1$ then $h'$ is splittable.
% Since $h_1$ and $h_3$ are consistent and $h'$ is splittable then Lemma \ref{consistent} shows that $h'$ is consistent.
% As $h'$ is complete and consistent it is a solution.
% Since $h_2$ specialises $h_1$ then $size(h_1) < size(h_2)$.
% Therefore, $size(h') < size(h_o)$ so $h_o$ is not an optimal solution and case (ii) cannot hold.
% These cases are exhaustive so the assumption cannot hold.
\end{proof}
% \begin{example}[\textbf{Specialisation constraint when fp=0}]
% % [\textbf{Specialisations of consistent programs}] 
% Consider the hypothesis:
% \[
%     \begin{array}{l}
%     h_1 = \left\{
%     \begin{array}{l}
%     \emph{f(A,B) $\leftarrow$ head(A,B)}
%     \end{array}
%     \right\}
%     \end{array}
% \]
% If $h_1$ does not entail any negative example, we can prune its specialisations, such as $h_2$ as they cannot appear in a rule of an optimal solution:
% \[
%     \begin{array}{l}
%     h_2 = \left\{
%     \begin{array}{l}
%     \emph{f(A,B) $\leftarrow$ head(A,B), even(B)}
%     \end{array}
%     \right\}
%     \end{array}
% \]

% \noindent
% It is important to note that case (a) does not prohibit specialisations of $h_1$ from being in multi-clause non-splittable programs such as:
% \[
%     \begin{array}{l}
%     h_3 = \left\{
%     \begin{array}{l}
%     \emph{f(A,B) $\leftarrow$ head(A,B),odd(B)}\\
%     \emph{f(A,B) $\leftarrow$ tail(A,C), f(C,B)}\\
%     \end{array}
%     \right\}
%     \end{array}
% \]
% \end{example}

\noindent
If a hypothesis does not entail any negative example, \name{} prunes all its specialisations, as they cannot be in a conjunction in an optimal solution:
\begin{corollary}
% [\textbf{Specialisation constraint when fp=0}]
Let $h_1$ be a hypothesis that does not entail any negative example and $h_2$ be a specialisation of $h_1$.
Then $h_2$ cannot be in a conjunction in an optimal solution.
\label{cor_consistent}
\end{corollary}
\begin{proof}
Follows from Proposition \ref{prop:consistent_appendix}.
\end{proof}

% Constraints inferred in cases (a), (b), and (c) never prune optimal solutions from the hypothesis space.
\noindent
We now show that the constrain stage never prunes an optimal solution:

\begin{proposition}[\textbf{Optimal soundness of the constraints}]
    % [\textbf{Optimally sound constraints}]
\label{prop_opt_sound}
The constrain stage of \name{} never prunes an optimal solution.
\end{proposition}
\begin{proof}
\name{} builds constraints from non-splittable non-separable programs that do not entail (a) any positive example, or (b) any negative example.
For case (a), if a program does not entail any positive example, \name{} prunes its specialisations which by Corollary \ref{cor_totincomplete} cannot be in a conjunction in an optimal solution.
For case (b), if a program does not entail any negative example, \name{} prunes its specialisations which by Corollary \ref{cor_consistent} cannot be in a conjunction in an optimal solution.
% Moreover, \name{} does not build subsumed conjunctions, which by Proposition \ref{prop_subsumed} cannot be in an optimal solution.
Therefore, \name{} never prunes an optimal solution.
\end{proof}

\subsection{\name{} Correctness}

We show the correctness of the \name{} without the constrain stage:
\begin{proposition}[\textbf{Correctness of \name{} without the constrain stage}]
\name{} without the constrain stage returns an optimal solution.
\label{correctness}
\end{proposition}
\begin{proof}
Proposition \ref{lem_generate} shows that \name{} can generate and test all non-splittable non-separable programs.
These generated programs include all joinable programs and all non-splittable combinable programs.
% \ac{do not understand: Among which are (i) all joinable programs, and (ii) all non-splittable combinable programs.}
% Proposition \ref{lem_join} shows that given all joinable programs, for any splittable combinable program, the join stage returns a conjunction with the same logical consequences. 
Proposition \ref{lem_join} shows that given all joinable programs, the join stage returns all non-subsumed splittable combinable programs.
% Proposition \ref{lem_join} shows that given all non-splittable non-separable programs, the join stage returns all non-subsumed non-separable combinable programs \ac{NEW}.
Therefore, \name{} builds all non-subsumed splittable and non-splittable combinable programs.
% non-subsumed combinable programs.
% there are finitely many splittable combinable programs and that the join stage can build them all \ac{from a quick read, is this true?}
% Given the set of all combinable programs, the combine stage returns a solution \cite{combo}.
% By Assumption \ref{existencesol}, a solution exists.
By Proposition \ref{prop_combine}, given all non-subsumed combinable programs, the combine stage returns an optimal solution.
Therefore, \name{} without the constrain stage returns an optimal solution.
\end{proof}

\noindent
We show the correctness of the \name{}:
\begin{theorem}[\textbf{Correctness of \name{}}]
\name{} returns an optimal solution.
\end{theorem}
\begin{proof}
From Proposition \ref{correctness}, \name{} without the constrain stage returns an optimal solution.
% \ac{can return every solution}.
From Proposition \ref{prop_opt_sound}, the constrain stage never prunes an optimal solution. 
Therefore, \name{} with the constrain stage returns an optimal solution.
% Moreover, \name{} uses iterative deepening over the program size to guarantee optimality. Therefore \name{} returns an optimal solution.
% \ac{From prop X, joiner without constraints always returns an optimal solution.
% From prop X, the constraint stage never prunes an optimal solution.
% Therefore, joiner with the constrain stage always returns an optimal solution.}
\end{proof}

\section{Experiments}
\label{sec:exp}

\subsection{Experimental domains}
We describe the characteristics of the domains and tasks used in our experiments.
% in Tables \ref{tab:dataset} and \ref{tab:tasks}. 
Figure \ref{fig:sols} shows example solutions for some of the tasks.

\paragraph{IGGP.}
In \emph{inductive general game playing} (IGGP)  \cite{iggp} the task is to induce a hypothesis to explain game traces from the general game playing competition \cite{ggp}.
IGGP is notoriously difficult for machine learning approaches.
The currently best-performing system can only learn perfect solutions for 40\% of the tasks.
Moreover, although seemingly a toy problem, IGGP is representative of many real-world problems, such as inducing semantics of programming languages \cite{DBLP:conf/ilp/BarthaC19}. 
We use 11 tasks: \emph{minimal decay (md)}, \emph{buttons-next}, \emph{buttons-goal}, \emph{rock - paper - scissors (rps)}, \emph{coins-next}, \emph{coins-goal}, \emph{attrition}, \emph{centipede}, \emph{sukoshi}, \emph{rainbow}, and \emph{horseshoe}.

\paragraph{Zendo.} Zendo is an inductive game in which one player, the Master, creates a rule for structures made of pieces with varying attributes to follow. The other players, the Students, try to discover the rule by building and studying structures which are labelled by the Master as following or breaking the rule. The first student to correctly state the rule wins. We learn four increasingly complex rules for structures made of at most 5 pieces of varying color, size, orientation and position. 
Zendo is a challenging game that has attracted much interest in cognitive science \cite{zendo}.

\paragraph{IMDB.}
The real-world IMDB dataset \cite{mihalkova2007} includes relations between movies, actors, directors, movie genre, and gender. It has been created from the International Movie Database (IMDB.com) database. We learn the relation \emph{workedunder/2}, a more complex variant \emph{workedwithsamegender/2}, and the disjunction of the two.
This dataset is frequently used \cite{alps}.

\subsection{Experimental Setup}
We measure the mean and standard error of the predictive accuracy and learning time.
%We enforce a timeout of 600s per task.
%We repeat all the experiments 20 times.
%We measure the mean and standard error.
We use a 3.8 GHz 8-Core Intel Core i7 with 32GB of ram.
The systems use a single CPU.

% \textbf{Q1.}
% We compare the performance of \popper{} and \name{} on all the tasks.
% We use \popper{} 2.0.0 \cite{popper2}.

% \textbf{Q2.}
% We compare the performance of \popper{} and \name{} when varying the size of the hypothesis space.
% We vary the maximum size of a rule allowed in a hypothesis, ie the maximum number of literals in a rule.
% We focus on the \emph{md} task.

\subsection{Experimental Results}

% \subsubsection{Comparison against other ILP systems}

We compare \name{} against \combo{} and \ale{}, which we describe below.

% \paragraph{\popper{}.}
% One key experimental question is to see whether \dcc{} can 
\begin{description}
% \item[\popper{}] 
% We compare against \popper{} as \name{} directly improves on it.
% For instance, if given only a single positive example, \name{} is identical to \popper{}.
% In all the experiments, we give \name{} and \popper{} identical biases.
\item[\combo{}] \combo{} uses identical biases to \name{} so the comparison is direct, i.e. fair.

\item[\ale{}] \ale{} excels at learning many large non-recursive rules and should excel at the trains and IGGP tasks.
Although \ale{} can learn recursive programs, it struggles to do so.
\name{} and \ale{} use similar biases so the comparison can be considered reasonably fair.

\end{description}
\noindent
We also tried/considered other ILP systems.
We considered \ilasp{} \cite{ilasp}.
However, \ilasp{} builds on \aspal{} and first precomputes every possible rule in a hypothesis space, which is infeasible for our datasets.
In addition, \ilasp{} cannot learn Prolog programs so is unusable in the synthesis tasks.
For instance, it would require precomputing $10^{15}$ rules for the \emph{coins} task.
% For instance, on the trains tasks, \ilasp{} took 2 seconds to pre-compute rules with three body literals; 20 seconds for rules with four body literals; and 12 minutes for rules with five body literals.  \ch{still doable in 1h timeout}
% Since the simplest train task requires rules with six body literals, \ilasp{} is unusable.

\begin{figure*}
\centering
\footnotesize
\begin{lstlisting}[caption=zendo1\label{zendo1}]
zendo1(A):- piece(A,C),size(C,B),blue(C),small(B),contact(C,D),red(D).
\end{lstlisting}
\centering

\begin{lstlisting}[caption=zendo2\label{zendo2}]
zendo2(A):- piece(A,B),piece(A,D),piece(A,C),green(D),red(B),blue(C).
zendo2(A):- piece(A,D),piece(A,B),coord1(B,C),green(D),lhs(B),coord1(D,C).
\end{lstlisting}
\centering

\begin{lstlisting}[caption=zendo3\label{zendo3}]
zendo3(A):- piece(A,D),blue(D),coord1(D,B),piece(A,C),coord1(C,B),red(C).
zendo3(A):- piece(A,D),contact(D,C),rhs(D),size(C,B),large(B).
zendo3(A):- piece(A,B),upright(B),contact(B,D),blue(D),size(D,C),large(C).
\end{lstlisting}
\centering

\begin{lstlisting}[caption=zendo4\label{zendo4}]
zendo4(A):- piece(A,C),contact(C,B),strange(B),upright(C).
zendo4(A):- piece(A,D),contact(D,C),coord2(C,B),coord2(D,B).
zendo4(A):- piece(A,D),contact(D,C),size(C,B),red(D),medium(B).
zendo4(A):- piece(A,D),blue(D),lhs(D),piece(A,C),size(C,B),small(B).\end{lstlisting}
\centering

\begin{lstlisting}[caption=pharma1\label{pharma1}]
active(A):- atm(A,B),typec(B),bond1(B,C),typeo(C),atm(A,F),types(F),bonddu(F,G),
            typen(G),atm(A,D),typeh(D),bond2(D,E),typedu(E).
\end{lstlisting}
\centering

\begin{lstlisting}[caption=pharma2\label{pharma2}]
active(A):- atm(A,B),types(C),bonddu(C,B),atm(A,G),typeh(F),bond2(F,G),
            atm(A,E),typec(D),bond1(D,E).
active(A):- atm(A,B),typena(C),bondv(B,C),atm(A,N),typen(O),bondu(N,O),
            atm(A,F),typedu(G),bondar(G,F).
\end{lstlisting}
\centering

\begin{lstlisting}[caption=pharma3\label{pharma3}]
active(A):- atm(A,D),bond2(D,E),typeh(E),atm(A,C),bond1(C,B),typec(B).
active(A):- atm(A,B),typena(C),bondv(B,C),atm(A,N),typen(O),bondu(N,O).
\end{lstlisting}
\centering

\begin{lstlisting}[caption=string2\label{string2}]
f_1(A):- head(A,B),cm(B).
f_1(A):- tail(A,B),f_1(B).
f_2(A):- head(A,B),cd(B).
f_2(A):- tail(A,B),f_2(B).
f_3(A):- head(A,B),co(B).
f_3(A):- tail(A,B),f_3(B).
f_4(A):- head(A,B),cr(B).
f_4(A):- tail(A,B),f_4(B).
f_5(A):- head(A,B),cu(B).
f_5(A):- tail(A,B),f_5(B).
f(A):- f_1(A),f_2(A),f_3(A),f_4(A),f_5(A).
\end{lstlisting}
\centering

% \begin{lstlisting}[caption=string3\label{string3}]
% f_1(A):- head(A,B),cl(B).
% f_1(A):- tail(A,B),f_1(B).
% f_2(A):- head(A,B),ch(B).
% f_2(A):- tail(A,B),f_2(B).
% f_3(A):- head(A,B),cj(B).
% f_3(A):- tail(A,B),f_3(B).
% f_4(A):- head(A,B),cm(B).
% f_4(A):- tail(A,B),f_4(B).
% f(A):- f_1(A),f_2(A).
% f(A):- f_3(A),f_4(A).
% \end{lstlisting}
\caption{Example solutions.}
\label{fig:sols}
\end{figure*}

\setcounter{figure}{7}
\begin{figure*}
\begin{lstlisting}[caption=minimal decay\label{minimaldecay}]
next_value(A,B):-c_player(D),c_pressButton(C),c5(B),does(A,D,C).
next_value(A,B):-c_player(C),my_true_value(A,E),does(A,C,D),my_succ(B,E),c_noop(D).
\end{lstlisting}

\begin{lstlisting}[caption=rainbow\label{rainbow}]
terminal(A):- mypos_r3(M),true_color(A,M,L),hue(L),mypos_r6(H),true_color(A,H,I),
              hue(I),mypos_r5(F),true_color(A,F,G),hue(G),mypos_r2(J),
              true_color(A,J,K),hue(K),mypos_r1(D),true_color(A,D,E),
              hue(E),mypos_r4(B),true_color(A,B,C),hue(C).
\end{lstlisting}

\begin{lstlisting}[caption=horseshoe\label{horseshoe}]
terminal(A):- int_20(B),true_step(A,B).
terminal(A):- mypos_b(E),mypos_c(G),true_cell(A,E,F),true_cell(A,G,F),mypos_d(D),
              true_cell(A,D,C),true_cell(A,B,C),adjacent(B,D).
terminal(A):- mypos_c(B),mypos_a(H),true_cell(A,H,I),mypos_b(J),true_cell(A,J,I),
              mypos_e(E),true_cell(A,B,D),true_cell(A,E,F),true_cell(A,G,F),
              adjacent(E,G),true_cell(A,C,D),adjacent(C,B).
\end{lstlisting}

\begin{lstlisting}[caption=buttons \label{buttons}]
next(A,B):-c_p(B),c_c(C),does(A,D,C),my_true(A,B),my_input(D,C).
next(A,B):-my_input(C,E),c_p(D),my_true(A,D),c_b(E),does(A,C,E),c_q(B).
next(A,B):-my_input(C,D),not_my_true(A,B),does(A,C,D),c_p(B),c_a(D).
next(A,B):-c_a(C),does(A,D,C),my_true(A,B),c_q(B),my_input(D,C).
next(A,B):-my_input(C,E),c_p(B),my_true(A,D),c_b(E),does(A,C,E),c_q(D).
next(A,B):-c_c(D),my_true(A,C),c_r(B),role(E),does(A,E,D),c_q(C).
next(A,B):-my_true(A,C),my_succ(C,B).
next(A,B):-my_input(C,D),does(A,C,D),my_true(A,B),c_r(B),c_b(D).
next(A,B):-my_input(C,D),does(A,C,D),my_true(A,B),c_r(B),c_a(D).
next(A,B):-my_true(A,E),c_c(C),does(A,D,C),c_q(B),c_r(E),my_input(D,C).
\end{lstlisting}
\begin{lstlisting}[caption=rps\label{rps}]
next_score(A,B,C):-does(A,B,E),different(G,B),my_true_score(A,B,F),beats(E,D),
                   my_succ(F,C),does(A,G,D).
next_score(A,B,C):-different(G,B),beats(D,F),my_true_score(A,E,C),
                   does(A,G,D),does(A,E,F).
next_score(A,B,C):-my_true_score(A,B,C),does(A,B,D),does(A,E,D),different(E,B).
\end{lstlisting}

\begin{lstlisting}[caption=coins\label{coins}]
next_cell(A,B,C):-does_jump(A,E,F,D),role(E),different(B,D),
                  my_true_cell(A,B,C),different(F,B).
next_cell(A,B,C):-my_pos(E),role(D),c_zerocoins(C),does_jump(A,D,B,E).
next_cell(A,B,C):-role(D),does_jump(A,D,E,B),c_twocoins(C),different(B,E).
next_cell(A,B,C):-does_jump(A,F,E,D),role(F),my_succ(E,B),
                  my_true_cell(A,B,C),different(E,D).
\end{lstlisting}

\begin{lstlisting}[caption=coins-goal\label{coins-goal}]
goal(A,B,C):- role(B),pos_5(D),my_true_step(A,D),score_100(C).
goal(A,B,C):- c_onecoin(E),my_pos(D),my_true_cell(A,D,E),score_0(C),role(B).
\end{lstlisting}
\caption{Example solutions.}
\end{figure*}

\section{Experimental results}

\subsection{Impact of the features}
Tables \ref{tab:accuracies_60} and \ref{tab:accuracies_600} show the results with 60s and 600s timeout respectively.

\begin{table}[ht]
% \small
\centering
\footnotesize
\begin{tabular}{@{}l|ccc@{}}
% \toprule
\multirow{2}{*}{\textbf{Task}} & \multirow{2}{*}{\textbf{\combo{}}} & \textbf{with join} & \multirow{2}{*}{\textbf{\name{}}}\\
& & \textbf{splittable} & \\
\midrule
\emph{attrition} & 67 $\pm$ 0 & \textbf{75 $\pm$ 0} & 67 $\pm$ 0\\
\emph{buttons} & 91 $\pm$ 0 & \textbf{100 $\pm$ 0} & \textbf{100 $\pm$ 0}\\
\emph{buttons-g} & \textbf{100 $\pm$ 0} & 90 $\pm$ 0 & 98 $\pm$ 2\\
\emph{centipede} & \textbf{100 $\pm$ 0} & \textbf{100 $\pm$ 0} & \textbf{100 $\pm$ 0}\\
\emph{coins} & 62 $\pm$ 0 & \textbf{100 $\pm$ 0} & 80 $\pm$ 0\\
\emph{coins-g} & \textbf{97 $\pm$ 0} & \textbf{97 $\pm$ 0} & \textbf{97 $\pm$ 0}\\
\emph{horseshoe} & 67 $\pm$ 0 & 69 $\pm$ 8 & \textbf{100 $\pm$ 0}\\
\emph{md} & \textbf{100 $\pm$ 0} & \textbf{100 $\pm$ 0} & \textbf{100 $\pm$ 0}\\
\emph{rainbow} & 50 $\pm$ 0 & \textbf{100 $\pm$ 0} & \textbf{100 $\pm$ 0}\\
\emph{rps} & 50 $\pm$ 0 & 82 $\pm$ 0 & \textbf{100 $\pm$ 0}\\
\emph{sukoshi} & 50 $\pm$ 0 & 50 $\pm$ 0 & \textbf{99 $\pm$ 0}\\\midrule
\emph{imdb1} & \textbf{100 $\pm$ 0} & \textbf{100 $\pm$ 0} & \textbf{100 $\pm$ 0}\\
\emph{imdb2} & \textbf{100 $\pm$ 0} & \textbf{100 $\pm$ 0} & \textbf{100 $\pm$ 0}\\
\emph{imdb3} & \textbf{100 $\pm$ 0} & \textbf{100 $\pm$ 0} & \textbf{100 $\pm$ 0}\\\midrule
\emph{pharma1} & 50 $\pm$ 0 & 56 $\pm$ 2 & \textbf{57 $\pm$ 1}\\
\emph{pharma2} & 50 $\pm$ 0 & \textbf{99 $\pm$ 1} & 98 $\pm$ 1\\
\emph{pharma3} & 50 $\pm$ 0 & \textbf{97 $\pm$ 1} & 96 $\pm$ 1\\\midrule
\emph{zendo1} & 94 $\pm$ 3 & \textbf{99 $\pm$ 0} & \textbf{99 $\pm$ 1}\\
\emph{zendo2} & 70 $\pm$ 1 & 92 $\pm$ 1 & \textbf{93 $\pm$ 1}\\
\emph{zendo3} & 79 $\pm$ 1 & 79 $\pm$ 1 & \textbf{80 $\pm$ 1}\\
\emph{zendo4} & 93 $\pm$ 1 & \textbf{99 $\pm$ 1} & \textbf{99 $\pm$ 1}\\\midrule
\emph{string1} & 50 $\pm$ 0 & \textbf{53 $\pm$ 1} & \textbf{53 $\pm$ 1}\\
\emph{string2} & 50 $\pm$ 0 & \textbf{53 $\pm$ 1} & \textbf{53 $\pm$ 1}\\
\emph{string3} & 50 $\pm$ 0 & 51 $\pm$ 2 & \textbf{100 $\pm$ 0}\\\midrule
\emph{denoising$\_$1c} & 52 $\pm$ 2 & \textbf{100 $\pm$ 0} & \textbf{100 $\pm$ 0}\\
\emph{denoising$\_$mc} & 60 $\pm$ 10 & \textbf{99 $\pm$ 1} & \textbf{99 $\pm$ 1}\\
\emph{fill} & 52 $\pm$ 2 & \textbf{100 $\pm$ 0} & \textbf{100 $\pm$ 0}\\
\emph{flip} & 50 $\pm$ 0 & \textbf{85 $\pm$ 9} & \textbf{85 $\pm$ 9}\\
\emph{hollow} & \textbf{55 $\pm$ 5} & \textbf{55 $\pm$ 5} & \textbf{55 $\pm$ 5}\\
\emph{mirror} & 57 $\pm$ 1 & \textbf{61 $\pm$ 4} & \textbf{61 $\pm$ 4}\\
\emph{move$\_$1p} & 50 $\pm$ 0 & \textbf{100 $\pm$ 0} & \textbf{100 $\pm$ 0}\\
\emph{move$\_$2p} & 50 $\pm$ 0 & 80 $\pm$ 8 & \textbf{88 $\pm$ 4}\\
\emph{move$\_$2p$\_$dp} & 55 $\pm$ 2 & 87 $\pm$ 8 & \textbf{88 $\pm$ 9}\\
\emph{move$\_$3p} & 50 $\pm$ 0 & 62 $\pm$ 9 & \textbf{64 $\pm$ 9}\\
\emph{move$\_$dp} & 55 $\pm$ 2 & \textbf{65 $\pm$ 6} & 62 $\pm$ 3\\
\emph{padded$\_$fill} & 52 $\pm$ 2 & \textbf{66 $\pm$ 3} & \textbf{66 $\pm$ 3}\\
\emph{pcopy$\_$1c} & 50 $\pm$ 0 & \textbf{86 $\pm$ 9} & 84 $\pm$ 9\\
\emph{pcopy$\_$mc} & 68 $\pm$ 6 & \textbf{97 $\pm$ 2} & \textbf{97 $\pm$ 2}\\
\emph{recolor$\_$cmp} & 51 $\pm$ 1 & \textbf{55 $\pm$ 4} & \textbf{55 $\pm$ 4}\\
\emph{recolor$\_$cnt} & 50 $\pm$ 0 & \textbf{51 $\pm$ 1} & \textbf{51 $\pm$ 1}\\
\emph{recolor$\_$oe} & 50 $\pm$ 0 & \textbf{52 $\pm$ 2} & \textbf{52 $\pm$ 2}\\
\emph{scale$\_$dp} & 61 $\pm$ 8 & 96 $\pm$ 2 & \textbf{97 $\pm$ 2}\\
\end{tabular}
\caption{
Predictive accuracies with a 60s timeout. We round accuracies to integer values. The error is standard error.
}
\label{tab:accuracies_60}
\end{table}

\begin{table}[ht]
% \small
\centering
\footnotesize
\begin{tabular}{@{}l|ccc@{}}
% \toprule
\multirow{2}{*}{\textbf{Task}} & \multirow{2}{*}{\textbf{\combo{}}} & \textbf{with join} & \multirow{2}{*}{\textbf{\name{}}}\\
& & \textbf{splittable} & \\
\midrule
\emph{attrition} & 67 $\pm$ 0 & \textbf{73 $\pm$ 2} & 67 $\pm$ 0\\
\emph{buttons} & \textbf{100 $\pm$ 0} & \textbf{100 $\pm$ 0} & \textbf{100 $\pm$ 0}\\
\emph{buttons-g} & \textbf{100 $\pm$ 0} & \textbf{100 $\pm$ 0} & 96 $\pm$ 2\\
\emph{centipede} & \textbf{100 $\pm$ 0} & \textbf{100 $\pm$ 0} & \textbf{100 $\pm$ 0}\\
\emph{coins} & 85 $\pm$ 9 & \textbf{100 $\pm$ 0} & \textbf{100 $\pm$ 0}\\
\emph{coins-g} & \textbf{97 $\pm$ 0} & \textbf{97 $\pm$ 0} & \textbf{97 $\pm$ 0}\\
\emph{horseshoe} & 67 $\pm$ 0 & \textbf{100 $\pm$ 0} & \textbf{100 $\pm$ 0}\\
\emph{md} & \textbf{100 $\pm$ 0} & \textbf{100 $\pm$ 0} & \textbf{100 $\pm$ 0}\\
\emph{rainbow} & 52 $\pm$ 0 & \textbf{100 $\pm$ 0} & \textbf{100 $\pm$ 0}\\
\emph{rps} & 77 $\pm$ 0 & \textbf{100 $\pm$ 0} & \textbf{100 $\pm$ 0}\\
\emph{sukoshi} & \textbf{100 $\pm$ 0} & \textbf{100 $\pm$ 0} & \textbf{100 $\pm$ 0}\\\midrule
\emph{imdb1} & \textbf{100 $\pm$ 0} & \textbf{100 $\pm$ 0} & \textbf{100 $\pm$ 0}\\
\emph{imdb2} & \textbf{100 $\pm$ 0} & \textbf{100 $\pm$ 0} & \textbf{100 $\pm$ 0}\\
\emph{imdb3} & \textbf{100 $\pm$ 0} & \textbf{100 $\pm$ 0} & \textbf{100 $\pm$ 0}\\\midrule
\emph{pharma1} & 50 $\pm$ 0 & \textbf{100 $\pm$ 0} & \textbf{100 $\pm$ 0}\\
\emph{pharma2} & 50 $\pm$ 0 & \textbf{98 $\pm$ 1} & 97 $\pm$ 1\\
\emph{pharma3} & 58 $\pm$ 5 & \textbf{96 $\pm$ 1} & \textbf{96 $\pm$ 1}\\\midrule
\emph{zendo1} & \textbf{100 $\pm$ 0} & \textbf{100 $\pm$ 0} & \textbf{100 $\pm$ 0}\\
\emph{zendo2} & 72 $\pm$ 1 & 93 $\pm$ 1 & \textbf{100 $\pm$ 0}\\
\emph{zendo3} & \textbf{79 $\pm$ 1} & 78 $\pm$ 1 & \textbf{79 $\pm$ 1}\\
\emph{zendo4} & 92 $\pm$ 1 & \textbf{98 $\pm$ 1} & \textbf{98 $\pm$ 1}\\\midrule
\emph{string1} & 50 $\pm$ 0 & 56 $\pm$ 1 & \textbf{100 $\pm$ 0}\\
\emph{string2} & 51 $\pm$ 0 & 56 $\pm$ 1 & \textbf{100 $\pm$ 0}\\
\emph{string3} & 49 $\pm$ 0 & \textbf{100 $\pm$ 0} & \textbf{100 $\pm$ 0}\\\midrule
\emph{denoising$\_$1c} & 52 $\pm$ 2 & \textbf{100 $\pm$ 0} & \textbf{100 $\pm$ 0}\\
\emph{denoising$\_$mc} & 60 $\pm$ 10 & \textbf{99 $\pm$ 1} & 97 $\pm$ 2\\
\emph{fill} & 59 $\pm$ 5 & \textbf{100 $\pm$ 0} & \textbf{100 $\pm$ 0}\\
\emph{flip} & 58 $\pm$ 8 & 94 $\pm$ 4 & \textbf{96 $\pm$ 4}\\
\emph{hollow} & 55 $\pm$ 5 & 85 $\pm$ 10 & \textbf{90 $\pm$ 6}\\
\emph{mirror} & 57 $\pm$ 1 & \textbf{78 $\pm$ 10} & 70 $\pm$ 3\\
\emph{move$\_$1p} & \textbf{100 $\pm$ 0} & \textbf{100 $\pm$ 0} & \textbf{100 $\pm$ 0}\\
\emph{move$\_$2p} & 50 $\pm$ 0 & 81 $\pm$ 9 & \textbf{91 $\pm$ 9}\\
\emph{move$\_$2p$\_$dp} & 55 $\pm$ 2 & 89 $\pm$ 9 & \textbf{100 $\pm$ 0}\\
\emph{move$\_$3p} & 50 $\pm$ 0 & 64 $\pm$ 9 & \textbf{85 $\pm$ 3}\\
\emph{move$\_$dp} & 55 $\pm$ 2 & 58 $\pm$ 2 & \textbf{83 $\pm$ 5}\\
\emph{padded$\_$fill} & 52 $\pm$ 2 & 66 $\pm$ 3 & \textbf{86 $\pm$ 5}\\
\emph{pcopy$\_$1c} & 50 $\pm$ 0 & 84 $\pm$ 9 & \textbf{92 $\pm$ 6}\\
\emph{pcopy$\_$mc} & 68 $\pm$ 6 & \textbf{97 $\pm$ 2} & 95 $\pm$ 3\\
\emph{recolor$\_$cmp} & 51 $\pm$ 1 & 58 $\pm$ 5 & \textbf{68 $\pm$ 5}\\
\emph{recolor$\_$cnt} & 50 $\pm$ 0 & 61 $\pm$ 7 & \textbf{80 $\pm$ 4}\\
\emph{recolor$\_$oe} & 50 $\pm$ 0 & 69 $\pm$ 8 & \textbf{75 $\pm$ 7}\\
\emph{scale$\_$dp} & 61 $\pm$ 8 & 96 $\pm$ 2 & \textbf{100 $\pm$ 0}\\
\end{tabular}
\caption{
Predictive accuracies with a 600s timeout. We round accuracies to integer values. The error is standard error.
}
\label{tab:accuracies_600}
\end{table}

\subsection{Comparison against other systems}
Tables \ref{tab:acc_sota_60} and \ref{tab:acc_sota_600} show the results with 60s and 600s timeout respectively.

\begin{table}[ht]
% \small
\centering
\footnotesize
\begin{tabular}{@{}l|cccc@{}}
\textbf{Task} & \textbf{\ale} & \textbf{\combo} & \textbf{\name}\\ \midrule
\emph{attrition} & 50 $\pm$ 0 & \textbf{67 $\pm$ 0} & \textbf{67 $\pm$ 0}\\
\emph{buttons} & 50 $\pm$ 0 & 91 $\pm$ 0 & \textbf{100 $\pm$ 0}\\
\emph{buttons-g} & 50 $\pm$ 0 & \textbf{100 $\pm$ 0} & 98 $\pm$ 2\\
\emph{centipede} & \textbf{100 $\pm$ 0} & \textbf{100 $\pm$ 0} & \textbf{100 $\pm$ 0}\\
\emph{coins} & 50 $\pm$ 0 & 62 $\pm$ 0 & \textbf{80 $\pm$ 0}\\
\emph{coins-g} & \textbf{100 $\pm$ 0} & 97 $\pm$ 0 & 97 $\pm$ 0\\
\emph{horseshoe} & 50 $\pm$ 0 & 67 $\pm$ 0 & \textbf{100 $\pm$ 0}\\
\emph{md} & 97 $\pm$ 0 & \textbf{100 $\pm$ 0} & \textbf{100 $\pm$ 0}\\
\emph{rainbow} & 50 $\pm$ 0 & 50 $\pm$ 0 & \textbf{100 $\pm$ 0}\\
\emph{rps} & 50 $\pm$ 0 & 50 $\pm$ 0 & \textbf{100 $\pm$ 0}\\
\emph{sukoshi} & 50 $\pm$ 0 & 50 $\pm$ 0 & \textbf{99 $\pm$ 0}\\\midrule
\emph{imdb1} & 60 $\pm$ 10 & \textbf{100 $\pm$ 0} & \textbf{100 $\pm$ 0}\\
\emph{imdb2} & 50 $\pm$ 0 & \textbf{100 $\pm$ 0} & \textbf{100 $\pm$ 0}\\
\emph{imdb3} & 50 $\pm$ 0 & \textbf{100 $\pm$ 0} & \textbf{100 $\pm$ 0}\\\midrule
\emph{pharma1} & 50 $\pm$ 0 & 50 $\pm$ 0 & \textbf{57 $\pm$ 1}\\
\emph{pharma2} & 50 $\pm$ 0 & 50 $\pm$ 0 & \textbf{98 $\pm$ 1}\\
\emph{pharma3} & 50 $\pm$ 0 & 50 $\pm$ 0 & \textbf{96 $\pm$ 1}\\\midrule
\emph{zendo1} & \textbf{100 $\pm$ 0} & 94 $\pm$ 3 & 99 $\pm$ 1\\
\emph{zendo2} & \textbf{100 $\pm$ 0} & 70 $\pm$ 1 & 93 $\pm$ 1\\
\emph{zendo3} & \textbf{99 $\pm$ 0} & 79 $\pm$ 1 & 80 $\pm$ 1\\
\emph{zendo4} & \textbf{99 $\pm$ 1} & 93 $\pm$ 1 & \textbf{99 $\pm$ 1}\\\midrule
\emph{string1} & 50 $\pm$ 0 & 50 $\pm$ 0 & \textbf{53 $\pm$ 1}\\
\emph{string2} & 50 $\pm$ 0 & 50 $\pm$ 0 & \textbf{53 $\pm$ 1}\\
\emph{string3} & 50 $\pm$ 0 & 50 $\pm$ 0 & \textbf{100 $\pm$ 0}\\\midrule
\emph{denoising$\_$1c} & 50 $\pm$ 0 & 52 $\pm$ 2 & \textbf{100 $\pm$ 0}\\
\emph{denoising$\_$mc} & 50 $\pm$ 0 & 60 $\pm$ 10 & \textbf{99 $\pm$ 1}\\
\emph{fill} & 50 $\pm$ 0 & 52 $\pm$ 2 & \textbf{100 $\pm$ 0}\\
\emph{flip} & 50 $\pm$ 0 & 50 $\pm$ 0 & \textbf{85 $\pm$ 9}\\
\emph{hollow} & 50 $\pm$ 0 & \textbf{55 $\pm$ 5} & \textbf{55 $\pm$ 5}\\
\emph{mirror} & 50 $\pm$ 0 & 57 $\pm$ 1 & \textbf{61 $\pm$ 4}\\
\emph{move$\_$1p} & 50 $\pm$ 0 & 50 $\pm$ 0 & \textbf{100 $\pm$ 0}\\
\emph{move$\_$2p} & 50 $\pm$ 0 & 50 $\pm$ 0 & \textbf{88 $\pm$ 4}\\
\emph{move$\_$2p$\_$dp} & 50 $\pm$ 0 & 55 $\pm$ 2 & \textbf{88 $\pm$ 9}\\
\emph{move$\_$3p} & 50 $\pm$ 0 & 50 $\pm$ 0 & \textbf{64 $\pm$ 9}\\
\emph{move$\_$dp} & 50 $\pm$ 0 & 55 $\pm$ 2 & \textbf{62 $\pm$ 3}\\
\emph{padded$\_$fill} & 50 $\pm$ 0 & 52 $\pm$ 2 & \textbf{66 $\pm$ 3}\\
\emph{pcopy$\_$1c} & 50 $\pm$ 0 & 50 $\pm$ 0 & \textbf{84 $\pm$ 9}\\
\emph{pcopy$\_$mc} & 50 $\pm$ 0 & 68 $\pm$ 6 & \textbf{97 $\pm$ 2}\\
\emph{recolor$\_$cmp} & 50 $\pm$ 0 & 51 $\pm$ 1 & \textbf{55 $\pm$ 4}\\
\emph{recolor$\_$cnt} & 50 $\pm$ 0 & 50 $\pm$ 0 & \textbf{51 $\pm$ 1}\\
\emph{recolor$\_$oe} & 50 $\pm$ 0 & 50 $\pm$ 0 & \textbf{52 $\pm$ 2}\\
\emph{scale$\_$dp} & 50 $\pm$ 0 & 61 $\pm$ 8 & \textbf{97 $\pm$ 2}\\
\end{tabular}
\caption{
Predictive accuracies with a 60s timeout. We round accuracies to integer values. The error is standard error.
% The tasks below the double line require negation while the one above do not.
%\ac{captions should not be centered} 
}
\label{tab:acc_sota_60}
\end{table}

\begin{table}[ht]
% \small
\centering
\footnotesize
\begin{tabular}{@{}l|cccc@{}}
% \toprule
\textbf{Task} & \textbf{\ale} & \textbf{\combo} & \textbf{\name}\\ \midrule
\emph{attrition} & 50 $\pm$ 0 & \textbf{67 $\pm$ 0} & \textbf{67 $\pm$ 0}\\
\emph{buttons} & \textbf{100 $\pm$ 0} & \textbf{100 $\pm$ 0} & \textbf{100 $\pm$ 0}\\
\emph{buttons-g} & \textbf{100 $\pm$ 0} & \textbf{100 $\pm$ 0} & 96 $\pm$ 2\\
\emph{centipede} & \textbf{100 $\pm$ 0} & \textbf{100 $\pm$ 0} & \textbf{100 $\pm$ 0}\\
\emph{coins} & 50 $\pm$ 0 & 85 $\pm$ 9 & \textbf{100 $\pm$ 0}\\
\emph{coins-g} & \textbf{100 $\pm$ 0} & 97 $\pm$ 0 & 97 $\pm$ 0\\
\emph{horseshoe} & 65 $\pm$ 0 & 67 $\pm$ 0 & \textbf{100 $\pm$ 0}\\
\emph{md} & 97 $\pm$ 0 & \textbf{100 $\pm$ 0} & \textbf{100 $\pm$ 0}\\
\emph{rainbow} & 50 $\pm$ 0 & 52 $\pm$ 0 & \textbf{100 $\pm$ 0}\\
\emph{rps} & \textbf{100 $\pm$ 0} & 77 $\pm$ 0 & \textbf{100 $\pm$ 0}\\
\emph{sukoshi} & 50 $\pm$ 0 & \textbf{100 $\pm$ 0} & \textbf{100 $\pm$ 0}\\\midrule
\emph{imdb1} & \textbf{100 $\pm$ 0} & \textbf{100 $\pm$ 0} & \textbf{100 $\pm$ 0}\\
\emph{imdb2} & 50 $\pm$ 0 & \textbf{100 $\pm$ 0} & \textbf{100 $\pm$ 0}\\
\emph{imdb3} & 50 $\pm$ 0 & \textbf{100 $\pm$ 0} & \textbf{100 $\pm$ 0}\\\midrule
\emph{pharma1} & 50 $\pm$ 0 & 50 $\pm$ 0 & \textbf{100 $\pm$ 0}\\
\emph{pharma2} & 50 $\pm$ 0 & 50 $\pm$ 0 & \textbf{97 $\pm$ 1}\\
\emph{pharma3} & 50 $\pm$ 0 & 58 $\pm$ 5 & \textbf{96 $\pm$ 1}\\\midrule
\emph{zendo1} & \textbf{100 $\pm$ 0} & \textbf{100 $\pm$ 0} & \textbf{100 $\pm$ 0}\\
\emph{zendo2} & \textbf{100 $\pm$ 0} & 72 $\pm$ 1 & \textbf{100 $\pm$ 0}\\
\emph{zendo3} & \textbf{100 $\pm$ 0} & 79 $\pm$ 1 & 79 $\pm$ 1\\
\emph{zendo4} & \textbf{99 $\pm$ 0} & 92 $\pm$ 1 & 98 $\pm$ 1\\\midrule
\emph{string1} & 50 $\pm$ 0 & 50 $\pm$ 0 & \textbf{100 $\pm$ 0}\\
\emph{string2} & 50 $\pm$ 0 & 51 $\pm$ 0 & \textbf{100 $\pm$ 0}\\
\emph{string3} & 50 $\pm$ 0 & 49 $\pm$ 0 & \textbf{100 $\pm$ 0}\\\midrule
\emph{denoising$\_$1c} & 50 $\pm$ 0 & 52 $\pm$ 2 & \textbf{100 $\pm$ 0}\\
\emph{denoising$\_$mc} & 50 $\pm$ 0 & 60 $\pm$ 10 & \textbf{97 $\pm$ 2}\\
\emph{fill} & 50 $\pm$ 0 & 59 $\pm$ 5 & \textbf{100 $\pm$ 0}\\
\emph{flip} & 50 $\pm$ 0 & 58 $\pm$ 8 & \textbf{96 $\pm$ 4}\\
\emph{hollow} & 55 $\pm$ 5 & 55 $\pm$ 5 & \textbf{90 $\pm$ 6}\\
\emph{mirror} & 60 $\pm$ 8 & 57 $\pm$ 1 & \textbf{70 $\pm$ 3}\\
\emph{move$\_$1p} & 50 $\pm$ 0 & \textbf{100 $\pm$ 0} & \textbf{100 $\pm$ 0}\\
\emph{move$\_$2p} & 50 $\pm$ 0 & 50 $\pm$ 0 & \textbf{91 $\pm$ 9}\\
\emph{move$\_$2p$\_$dp} & 50 $\pm$ 0 & 55 $\pm$ 2 & \textbf{100 $\pm$ 0}\\
\emph{move$\_$3p} & 50 $\pm$ 0 & 50 $\pm$ 0 & \textbf{85 $\pm$ 3}\\
\emph{move$\_$dp} & 50 $\pm$ 0 & 55 $\pm$ 2 & \textbf{83 $\pm$ 5}\\
\emph{padded$\_$fill} & 50 $\pm$ 0 & 52 $\pm$ 2 & \textbf{86 $\pm$ 5}\\
\emph{pcopy$\_$1c} & 50 $\pm$ 0 & 50 $\pm$ 0 & \textbf{92 $\pm$ 6}\\
\emph{pcopy$\_$mc} & 50 $\pm$ 0 & 68 $\pm$ 6 & \textbf{95 $\pm$ 3}\\
\emph{recolor$\_$cmp} & 50 $\pm$ 0 & 51 $\pm$ 1 & \textbf{68 $\pm$ 5}\\
\emph{recolor$\_$cnt} & 56 $\pm$ 6 & 50 $\pm$ 0 & \textbf{80 $\pm$ 4}\\
\emph{recolor$\_$oe} & 50 $\pm$ 0 & 50 $\pm$ 0 & \textbf{75 $\pm$ 7}\\
\emph{scale$\_$dp} & 50 $\pm$ 0 & 61 $\pm$ 8 & \textbf{100 $\pm$ 0}\\
\end{tabular}
\caption{
Predictive accuracies with a 600s timeout. We round accuracies to integer values. The error is standard error.
% The tasks below the double line require negation while the one above do not.
%\ac{captions should not be centered} 
}
\label{tab:acc_sota_600}
\end{table}

\end{appendices}

\end{document}